\documentclass[lettersize,journal]{IEEEtran}
\usepackage{amsmath,amsfonts}
\usepackage{array}
\usepackage{graphicx} % 基础图片插入
\usepackage{subfigure} % 子图管理
\usepackage{textcomp}
\usepackage{stfloats}
\usepackage{url}
\usepackage{verbatim}
\usepackage{graphicx}
\usepackage{cite}
\usepackage[pagebackref,breaklinks,colorlinks]{hyperref}
\usepackage{bbm}
\usepackage{multirow}
\usepackage{makecell}
\usepackage{diagbox}
\usepackage{tablefootnote}
\usepackage{threeparttable}
\usepackage{booktabs}
\usepackage[cmyk]{xcolor}
\usepackage[ruled,linesnumbered]{algorithm2e}
\usepackage[switch]{lineno}

\hypersetup{
  colorlinks=true, % 用颜色区分链接，而非边框
  linkcolor=blue, % 正文引用链接颜色（如蓝色）
  citecolor=blue, % 文献引用链接颜色（如红色）
  urlcolor=black, % 网址链接颜色（如绿色）
}
\begin{document}

\title{Inter- and Intra-image Refinement for Few Shot Segmentation}
\author{Ourui Fu, Hangzhou He, Kaiwen Li, Xinliang Zhang, Lei Zhu, Shuang Zeng, Zhaoheng Xie, Yanye Lu
    \thanks{This work was supported in part by the National Natural Science Foundation of China under grants 82371112, 623B2001; in part by the Science Foundation of Peking University Cancer Hospital (JC202505); in part by the China National Postdoctoral Program for Innovative Talents (BX20250368). (Corresponding authors: Yanye Lu)}
    \thanks{Ourui Fu, Hangzhou He, Kaiwen Li, Xinliang Zhang, Lei Zhu, Shuang Zeng, Zhaoheng Xie, Yanye Lu are with the Department of Biomedical Engineering, Peking University, Beijing 100871, China; also with the Institute of Medical Technology, Peking University Health Science Center, Peking University, Beijing 100191, China; also with the National Biomedical Imaging Center, Peking University, Beijing 100871, China. (e-mail: orfu@stu.pku.edu.cn; zhuang@stu.pku.edu.cn; kaiwenli325@gmail.com; zhangxinliang\_mit@163.com; zhulei@pku.edu.cn; stevezs@pku.edu.cn; xiezhaoheng@pku.edu.cn; yanye.lu@pku.edu.cn)}
}
% \markboth{IEEE TRANSACTIONS ON IMAGE PROCESSING, VOL.X, NO.X, X 2025}%

% \linenumbers


\maketitle
\begin{abstract}
Deep neural networks for semantic segmentation rely on large-scale annotated datasets, leading to an annotation bottleneck that motivates few shot semantic segmentation (FSS) which aims to generalize to novel classes with minimal labeled exemplars. Most existing FSS methods adopt a prototype-based paradigm, which generates query prior map by extracting masked-area features from support images and then makes predictions guided by the prior map. However, they suffer from two critical limitations induced by inter- and intra-image discrepancies: 1) The intra-class gap between support and query images, caused by single-prototype representation, results in scattered and noisy prior maps; 2) The inter-class interference from visually similar but semantically distinct regions leads to inconsistent support-query feature matching and erroneous predictions. To address these issues, we propose the Inter- and Intra-image Refinement (IIR) model. The model contains an inter-image class activation mapping based method that generates two prototypes for class-consistent region matching, including core discriminative features and local specific features, and yields an accurate and robust prior map. For intra-image refinement, a directional dropout mechanism is introduced to mask inconsistent support-query feature pairs in cross attention, thereby enhancing decoder performance. Extensive experiments demonstrate that IIR achieves state-of-the-art performance on 9 benchmarks, covering standard FSS, part FSS, and cross-domain FSS. Our source code is available at \href{https://github.com/forypipi/IIR}{https://github.com/forypipi/IIR}.
\end{abstract}


\section{Introduction}
\label{sec:intro}
\IEEEPARstart{D}EEP neural networks have achieved remarkable success in various computer vision tasks, including image classification \cite{ResNet}, and semantic segmentation \cite{UNet, FCN}. However, dense prediction tasks such as semantic segmentation typically require large-scale and pixel-level annotated datasets for training, which involve substantial time and labor costs. Furthermore, conventional semantic segmentation models struggle to generalize to novel categories with only a few annotated samples. 
To address these limitations, researchers have extensively investigated few shot semantic segmentation (FSS) approaches \cite{SG-ONE, CGMGM, CyCTR, DCAMA, DCP, HSNet, IPRNet, HM, MIANet}.

\begin{figure}[t]
  \centering
  % 第一张子图：宽度0.45栏宽，子标题(a)
  \subfigure[Our model’s prior map highlights the proper foreground and contains less noise, resulting in more complete predictions within the red circle.]{
    \includegraphics[width=0.45\textwidth]{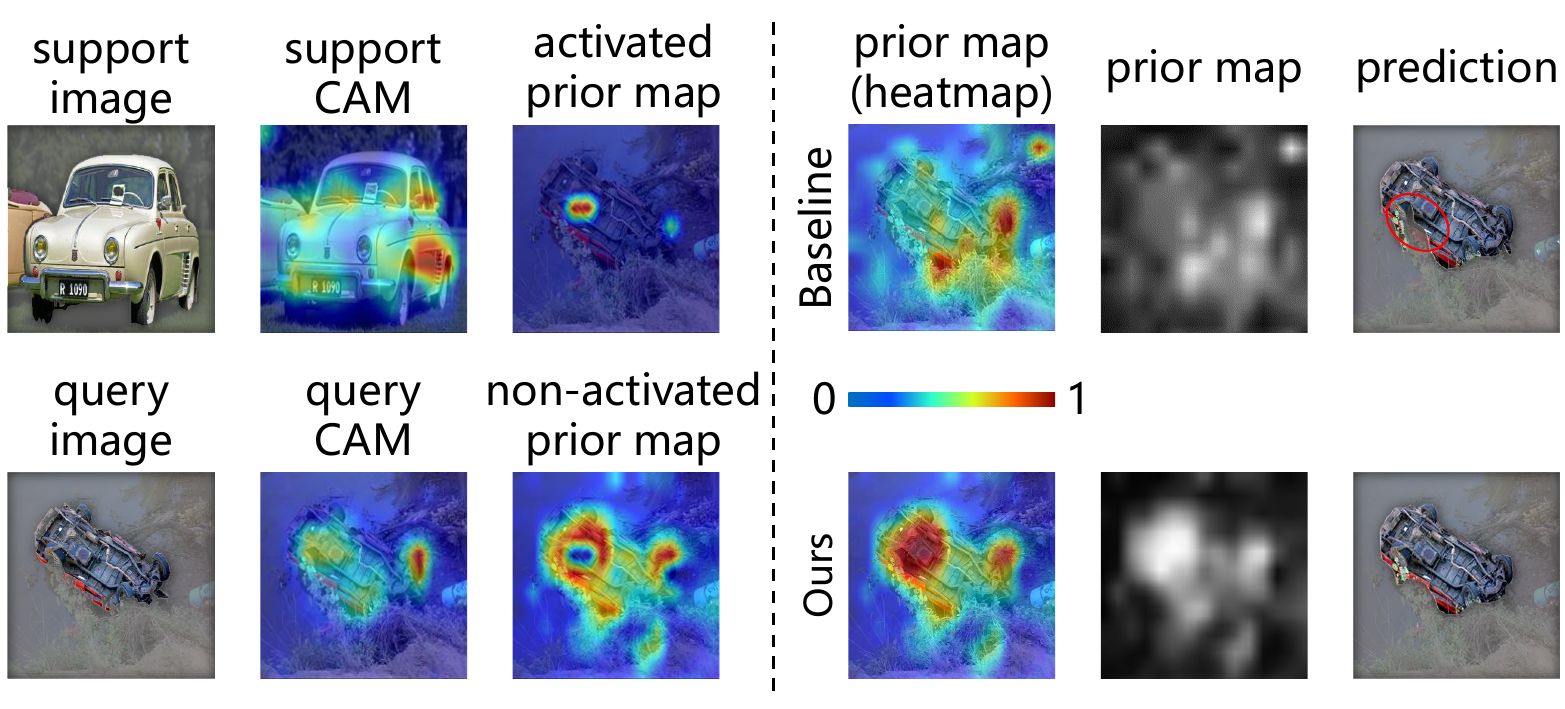}
    \label{subfig:inter} % 子图交叉引用标签
  }
  \hfill % 间距
  % 第二张子图：宽度0.45栏宽，子标题(b)
  \subfigure[The green and blue circles represent confusing background and foreground regions in the query and support images, respectively.]{
    \includegraphics[width=0.45\textwidth]{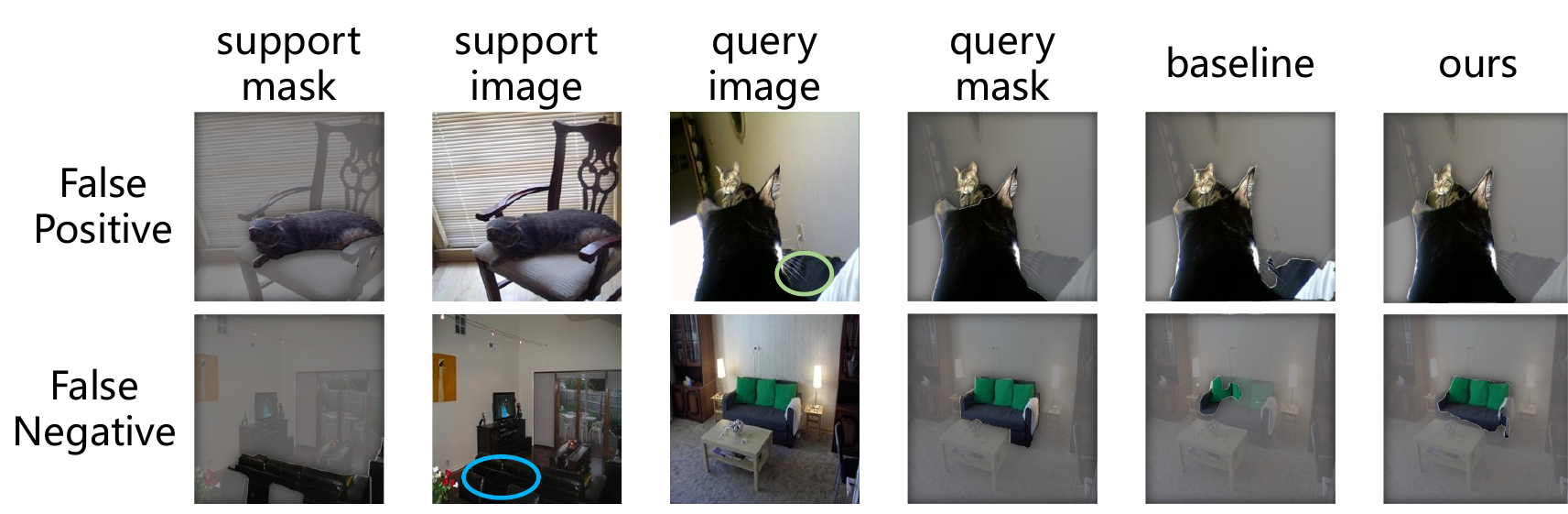}
    \label{subfig:intra} % 子图交叉引用标签
  }
  % 总标题
  \caption{(a) Inter-Image Gap: capturing foreground core features and local specific features enables a more precise and robust prior map; (b) Intra-Image Gap: masking inconsistent pixel pairs mitigates false positive and false negative predictions.}
  \label{fig:inter_intra}
\end{figure}

% \IEEEpubidadjcol
The predominant FSS methodology employs a prototype-based paradigm, where foreground features extracted from masked support images are aggregated into a class prototype to guide query segmentation via similarity matching. In practice, the feature encoder is often frozen, and only the segmentation decoder is optimized. Consequently, the final model performance critically depends on two core factors: the quality of the \textbf{Prior Map} and the efficacy of the \textbf{Decoder}. 
To improve prior map quality, works such as AENet \cite{AENet} and TBSNet \cite{TBS} focus on enhancing prototype representation by aggregating foreground and background features or introducing attention mechanisms. For decoder optimization, methods like CyCTR \cite{CyCTR} and SCCAN \cite{SCCAN} design cross-attention modules or feature fusion blocks to refine the prior map.

Despite these targeted efforts, most existing methods still fail to explicitly account for the inherent variations between and within support/query samples, overlooking the fundamental impact of inter- and intra-sample differences on both the prior map generation and support-query feature alignment.
Our study finds that inter-support-query and intra-support/intra-query image differences lead to inaccurate prior maps and support-query feature mismatch in decoder, respectively. These two issues collectively degrade the generalization ability of FSS models to novel categories, which motivates our research.

\begin{figure}[t]
  \centering
  \includegraphics[width=\linewidth]{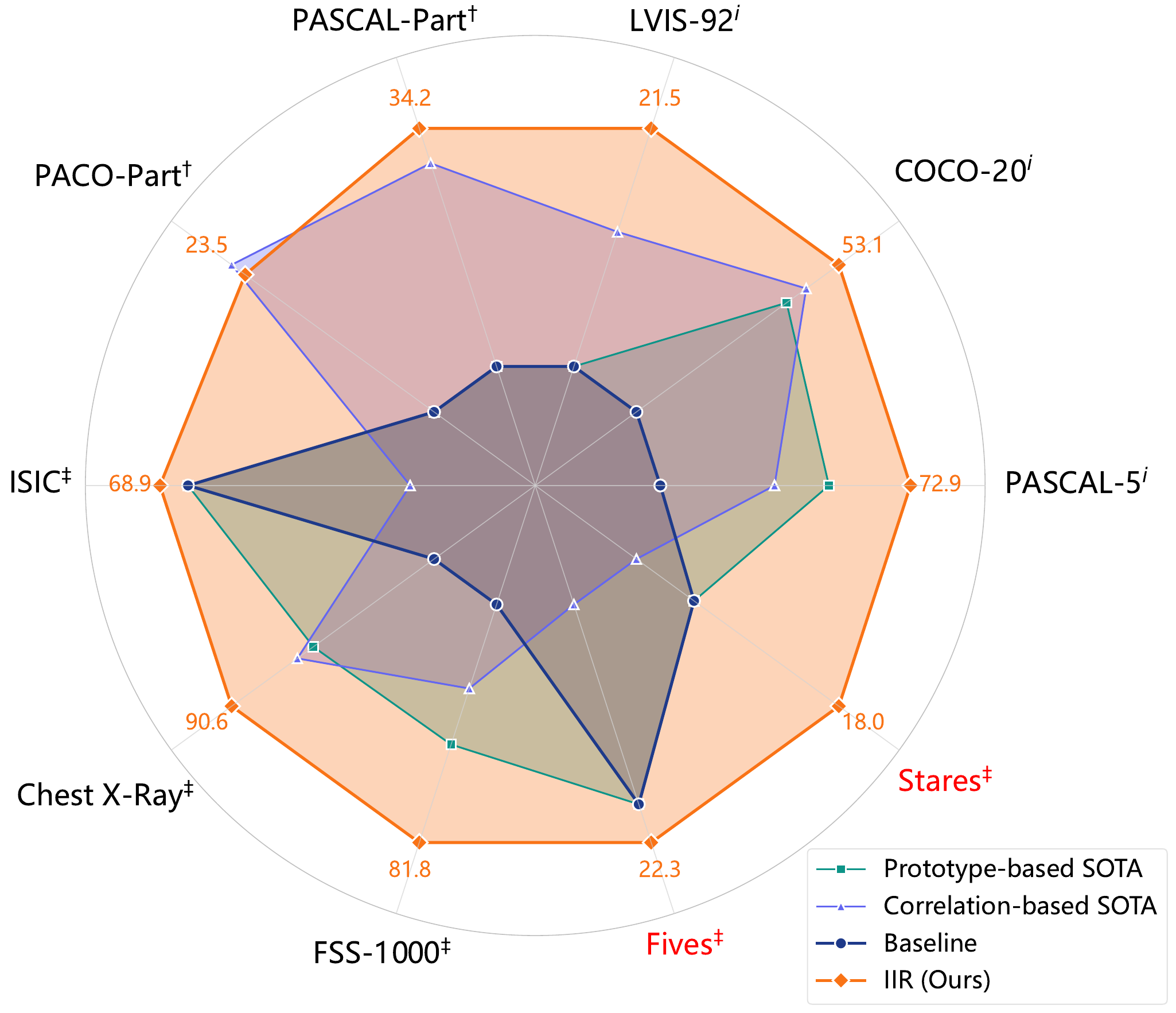}
  \caption{Our prototype-based method outperforms the baseline across 10 datasets, including standard FSS, object part FSS (with superscript $\dagger$), and cross domain FSS (with superscript $\ddagger$). The new challenging fundus vessel cross domain FSS datasets are highlighted in red.}
  \label{fig:radar}
\end{figure}

\textbf{Inter-Image Gap}, resulting from the use of a single prototype, will cause the prior map to be scattered and inaccurate (e.g., obvious background noise in the prior map of baseline models, as shown in Fig. \ref {subfig:inter} top-right). Specifically, using a single prototype is vulnerable to the internal heterogeneity of the foreground in two aspects:
\begin{enumerate}
    \item Support prototype lacks of diversity: the foreground region of the support image contains various details, like different colors and texture changes. Simply adopting a single masked average pooling for support prototype generation will obscure such fine-grained details, thereby compromising representational capacity for local foreground details in the query image; 
    \item Query image has interference: similarity is calculated between all the query image regions and the same support foreground prototype. Consequently, if there are background regions in the query image that are similar to the foreground prototype, they will be misjudged as foreground.
\end{enumerate}

To mitigate the limitations of generating a single prototype from a single area, we partition the support image and query image into core discriminative and local specific regions and extract prototypes therefrom. By performing similarity matching between each of these prototypes and the corresponding consistent regions in the query image, we generate a more accurate and robust prior map (Fig. \ref{subfig:inter}, bottom-right).

\textbf{Intra-Image Interference} stems from the mutual confusion between similar foreground and background within the same image, which in turn is reflected in the decoder cross attention map. This results in the most similar support-query pairs exhibiting inconsistent masks during training, which is referred to as support-query inconsistency.
More specifically, support-query inconsistency originates from two types of Intra-Image Interference, as illustrated in the Fig. \ref{subfig:intra}:
\begin{enumerate}
    \item False Positive: the query image itself contains similar background regions (e.g., in color, texture, etc.), and the decoder misclassifies the background as the foreground, as shown in Fig. \ref{subfig:intra} top; 
    \item False Negative: background interference in the query image also causes the decoder to misclassify the foreground of the query image as the background, as shown in Fig. \ref{subfig:intra} bottom.
\end{enumerate}

To tackle the issue of support-query inconsistency in the decoder, we first identify inconsistent support-query pairs via cross attention maps and then apply masking to these pairs during the training phase. By reducing support-query feature inconsistency, our method boosts decoder performance.

Building on the two aforementioned improvements, which target inter-image gaps via multi-prototype extraction and resolve intra-image interference via directional dropout, we further propose the \textbf{IIR} model (\textbf{I}nter- and \textbf{I}ntra-image \textbf{R}efinement), which simultaneously addresses both inter-image and intra-image discrepancies in FSS.
As shown in Fig. \ref{fig:radar}, comprehensive experiments on 10 datasets across three tasks (standard FSS, object part FSS and cross-domain FSS) validate the effectiveness of our proposed model in enhancing segmentation performance under diverse FSS scenarios.

We summarize our key contributions as follows:
\begin{itemize}
\item A CAM-based prior map generation method that addresses the inter-image discrepancy, yielding a more accurate and robust segmentation prior map.
\item An inconsistency masking approach that suppresses misleading feature matching pairs and intra-image interference, which contributes to refining the decoder’s feature alignment and prediction performance.
\item Our method achieves state-of-the-art performance on 9 benchmarks, including two newly created fundus vessel segmentation datasets, spanning standard, object part, and cross domain FSS tasks.
\end{itemize}

The paper is organized as follows: section \ref{sec:Related_Work} reviews the related work in recent years; section \ref{sec:Method} elaborates on the proposed method in detail; section \ref{sec:Exp} presents the experimental results and corresponding analysis; section \ref{sec:discussion} provides further discussions on our motivations and results; Finally, section \ref{sec:conclusion} summarizes the content of this paper, its limitations, and potential future research directions for improvement.
\section{Related Work}
\label{sec:Related_Work}

\subsection{Few Shot Learning (FSL)}
\label{subsec:FSL}
Most deep learning methods require a huge amount of data for training. However, collecting a large dataset is a time-consuming and labor-intensive process. To alleviate this burden, researchers focus on few shot learning (FSL) \cite{PrototypeNet, MatchingNet, Meta-LSTM, SiameseNet}, which aims to improve model generalization when trained with only a limited amount of annotated data.
PrototypeNet \cite{PrototypeNet} introduced the prototypical network which is widely used in FSL. Instead of comparing the features from all support images with the query features, the model calculates the average features of the support images as a class prototypical feature, then the query image is classified by comparing its feature with the class prototype.

\subsection{Few Shot Segmentation (FSS)}
\label{subsec:FSS}
Few shot segmentation (FSS) is an application of FSL in semantic segmentation tasks \cite{UNet, Deeplab}, aiming to assign each pixel to a semantic class under limited data constraints.
FSS models can be categorized into two main classes: prototypical-based methods and 4D correlation-based methods.
Prototypical-based methods \cite{IPRNet,MIANet,RARE,CGMGM,ABCB,AENet,DBMNet,PFENet,PFENet++,OCNet,PAHNet,DCP,HPA} concentrate on comparing the features of the query images with a prototype representing the foreground class in the support set. Instead of pixel-level comparisons, these methods leverage the concept of prototypical representation to guide the segmentation task.
However, most methods employ only a single prototype as the representative feature of the foreground class in the support image. Such prototype features are susceptible to gaps between the support and query images, leading to inactivated foreground regions or false activation of background regions, and thus resulting in a prior map with substantial noise. 
In contrast, we obtain two prototypes by leveraging class cues, which enhances the robustness of the prototype features and yields a prior map with more accurate activations, as shown in section \ref{sec:motivation PM}.

4D correlation-based methods \cite{HSNet,VAT,CyCTR,TBS,HM} can be summarized as follows: it performs pixel-wise matching between support images and query images to form a 4D tensor, and then decodes this tensor via 4D convolution to predict the mask of the query image.
CyCTR \cite{CyCTR} introduces a cyclic consistency constraint by masking the entire column of cross attention map when meets support→query→support inconsistent. However, 
our analysis reveals that this column-wise consistency over-masks more than 30\% pixels in average, leading to severe information loss. 
In response to this critical observation, we develop a pixel-wise consistency verification mechanism that masks unidirectional support→query inconsistencies, effectively reducing the over-mask ratio as discussed in section \ref{sec:motivation DD}.

In recent years, with the widespread application of SAM \cite{SAM, SAM2} in the field of semantic segmentation, a surge of works \cite{Matcher,GFSAM,DSV-LFS,FSSAM} has emerged to explore the segmentation capabilities of SAM under data-scarce scenarios. PerSAM \cite{PerSAM} focuses on the train-free paradigm: it selects positive and negative sample points with the highest and lowest similarity values from the prior map, and uses the prior map to guide the decoding process of the SAM decoder. This urges the prompt tokens to focus on foreground target regions, thereby enhancing feature aggregation. 
However, a series of SAM-based methods are all trained or evaluated on the pre-trained SAM backbone, which has been previously trained on the SA-1B large-scale dataset. Its ``segment anything" capability is fundamentally built on ``having seen everything", which constitutes a fundamental conflict with the requirement of ``generalization to unseen classes" in few shot segmentation.

\subsection{Cross Domain Few Shot Segmentation (CDFSS)}
PATNet \cite{PATNet} further extends few shot segmentation to cross domain few shot segmentation (CDFSS). This task rejects the assumption of few shot segmentation that training and testing domains are the same (e.g., in PASCAL-5$^i$, the training set categories include aeroplane, bicycle, bird, boat, and bottle, while the testing set categories include pottedplant, sheep, sofa, train, and tvmonitor—all belonging to the natural image domain) and replaces the testing domain with domains such as remote sensing domain and medical domain.
IFA \cite{IFA} proposes a fine-tuning strategy, which achieves model finetuning by treating augmented support images as pseudo query images.
Several models \cite{APM,LoEC,DFN} address the domain shift issue in cross domain scenarios by designing customized modules or optimization strategies, thereby achieving the adaptation and decoupling of domain-related information and further enhancing the models’ cross domain generalization performance.
\section{Method}
\label{sec:Method}
In this section, we first summarize FSS task definition and our IIR model in \ref{sec:Overview}. Then we present the Prior Map Generation Module (PMGM) in Section \ref{subsec:PMGM}, followed by discussion of the Directional Dropout Module (DDM) and the decoder in Section \ref{subsec:Decoder}.

\subsection{Problem and Method Overview}
\label{sec:Overview}
In general, our method follows the standard episodic training process of FSS. 
The dataset is divided into a training set ($\mathcal{D}_{train}$) and a test set ($\mathcal{D}_{test}$). The two sets contain non-overlapping foreground classes.
An episode is extracted from the dataset, composed of a support set $\mathcal{S}=\{I^s, M^s\}$ and a query set $\mathcal{Q}=\{I^q, M^q\}$, where $I$ and $M$ represent the image and its binary mask, respectively. 
During training, the model samples an episode from $\mathcal{D}_{train}$ and learns to map $(I^s, M^s, I^q)$ to $M^q$.
During testing, the model is frozen and episodes are sampled from $\mathcal{D}_{test}$ to predict $M^q$ based on $(I^s, M^s, I^q)$.

% 第一段：细化Baseline流程，呼应前文术语并点出局限
Given a support image-mask pair, the prototype-based paradigm first extracts a single foreground prototype $p^s\in\mathbb{R}^c$ through masked average pooling (MAP).
Next, the model generates the prior map $P\in\mathbb{R}^{1\times h\times w}$ by performing pixel-wise similarity matching between this single prototype and the whole feature map of the query image. This prior map serves as the initial guidance for subsequent segmentation.
For further feature refinement, the model employs a cross attention decoder. In this decoder, flatten support features $X^s\in\mathbb{R}^{hw\times c}$ are treated as \textit{keys/values} and flatten query features $X^q\in\mathbb{R}^{hw\times c}$ as \textit{queries} to facilitate feature interaction. Finally, the refined feature representation is fed into a segmentation head to get the final binary mask of the query image $\hat{M}^q$.

The standard prototype paradigm has two critical limitations. First, the single prototype generated by MAP fails to capture fine-grained foreground details, leading to the inter-image gap and resulting in inaccurate prior maps. Second, the cross attention module of the model lacks effective mechanisms to suppress misleading feature correspondences, which causes intra-image inconsistency between support and query mask alignments. To address these limitations, our method designs two targeted modules with an overview presented in Fig. \ref{fig:model_arch}:

\begin{figure*}[tbhp]
  \centering
  \includegraphics[width=\textwidth]{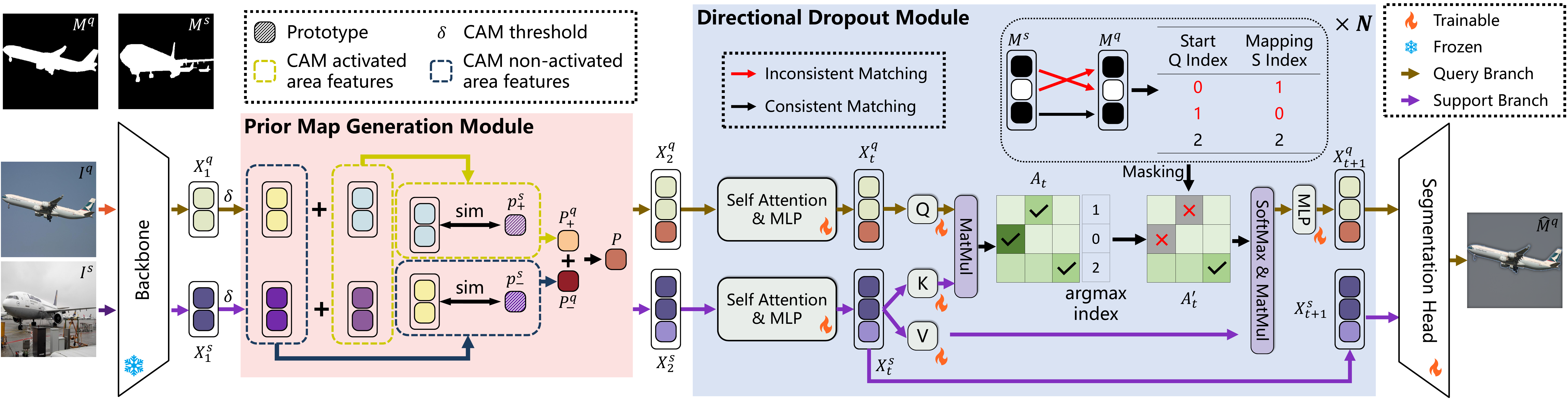}
  \caption{
  \textbf{Overall architecture of the proposed model.} Support (superscript notation \textit{s}) and query images (superscript notation \textit{q}) are firstly processed through a frozen backbone encoder to extract preliminary features $X^s_1$ and $X^q_1$, respectively. These features, combined with the support mask $M^s$ are subsequently fed into our Prior Map Generation Module (PMGM), which generates a CAM-based support and query prior map. Next, support branch guides query branch training through Directional Dropout Module (DDM).
  In DDM, the query feature serves as the \textit{Query}, while the support feature acts as the \textit{Key} and \textit{Value} in the cross attention. 
  The black and red line and text in DDM means the consistent and inconsistent support→query mapping. Dashed box in DDM is exclusively guide training procedure.
  }
  \label{fig:model_arch}
\end{figure*}

\begin{itemize}
    \item The Prior Map Generation Module (PMGM) aims to resolve the inter-image gap. Different from existing approaches that use MAP to produce a single prototype, the PMGM decomposes the support feature space into two complementary regions via Class Activation Mapping (CAM) heatmap. These regions are CAM activated regions that highlight discriminative cores and CAM non-activated regions that capture local details. 
    From these two regions combined with mask constraints,  PMGM extracts three specialized prototypes instead of one. These prototypes include an activated prototype $p^s_+\in\mathbb{R}^c$ which contains foreground core information, a non-activated prototype $p^s_-\in\mathbb{R}^c$ which integrates foreground local details, as well as the original prototype $p^s$ that retains the initial feature representation. By performing region-consistent similarity matching between each prototype and the corresponding CAM-aligned regions in the query image, PMGM outputs a more accurate and robust refined prior map, effectively mitigating the prior map inaccuracy of previous model.
    \item The Directional Dropout Module (DDM) targets the intra-image inconsistency of the previous model. For the decoder stage, our method adopts a dual-branch interaction mechanism. Support and query features are fed into each decoder branches. The support branch first refines the prior map generated by the PMGM to align with the support's ground-truth mask. This refined representation then guides the query branch to optimize mask prediction. During this guidance process, the DDM dynamically identifies and masks inconsistent support-query feature pairs in the cross attention map. The DDM prevents intra-image inconsistency from propagating to the final prediction, further enhancing decoder's performance.
\end{itemize}

During the testing phase, due to the absence of $M_q$, the model does not perform masking on the cross attention map; to avoid the inconsistency between the training and testing distributions, the model dynamically increases the values of unmasked pairs according to the masking ratio during the training phase.
For the K-shot setting, we input K support images into the model. PMGM average K prior maps and prototypes while decoders share a unified support branch.

\subsection{Prior Map Generation Module (PMGM)}
\label{subsec:PMGM}
To address the inter-image feature gap between support and query images and mitigate the inaccuracy of prior maps, the PMGM is specifically designed to refine prior map generation by leveraging class-discriminative region information.
Given a specific class $o^s$ from the support image, CAM method generates a heatmap that highlights the regions contributing significantly to classifying the image into that class.
PMGM leverages CAM to localize core discriminative (activated) and local specific (non-activated) regions, extracting two additional prototypes from these regions and catching class information to fill the gap between support and query features, thus reducing the noise and resulting in a more accurate prior map.

\subsubsection{Prototype Extraction}
To capture both core discriminative regions and local detail regions of the target class, we generate CAM heatmaps for support and query images.
Firstly, the query feature $X^q_1\in\mathbb{R}^{c\times h\times w}$ and the support feature $X^s_1\in\mathbb{R}^{c\times h\times w}$ are extracted from the frozen backbone. Since the support and query foreground share the same class, the class $o^s$ can be inferred by the high-level support foreground features $X^s_1$ and support mask $M^s\in\mathbb{R}^{1\times h\times w}$ with a classification head:
\begin{equation}
\begin{aligned}
  X^s_1=\mathcal{F}_{en}(I^s),X^q_1=\mathcal{F}_{en}(I^q),\\
  o^s=\arg\max(\mathcal{F}_{cls}(X^s_1\odot M^s)),
\label{eq:classification}
\end{aligned}
\end{equation}
where $\mathcal{F}_{en}$ and $\mathcal{F}_{cls}$ refer to the encoder and classification head, and $\odot$ represents element-wise multiplication. Based on the implicit $o^s$, support heatmap $C^s$ and query heatmap $C^q$ are generated to highlight the most discriminative object regions.

Based on $C^s$ and $C^q$, we further explicitly distinguish core discriminative regions and local detail regions. Four binary masks are generated based on a hyperparameter threshold $\delta$ and the indicator function$\mathbbm{1}(\cdot)$ (i.e.,$\mathbbm{1}(\text{true}) = 1$ and $\mathbbm{1}(\text{false}) = 0$). Specifically:
\begin{itemize}
    \item The support activated mask $C^s_+$ highlights core discriminative regions: $C^s_+=\mathbbm{1}(C^s\geq \delta)$;
    \item The query activated mask $C^q_+$ emphasizes core distinguishing regions: $C^q_+=\mathbbm{1}(C^q\geq \delta)$;
    \item The support non-activated mask $C^s_-$ capture local detail regions: $C^s_-=1-C^s_+$;
    \item The query non-activated mask $C^q_-$ identifies fine-grained local regions: $C^q_-=1 - C^q_+$.
\end{itemize}

For each foreground region, PMGM derives three prototypes $p^s_+, p^s_-$, and $p^s$ using MAP:
\begin{equation}
  p^s_i=\frac{\sum_{x,y}X^s_1(x,y)\mathbbm{1}(M^s(x,y) \wedge C^s_i(x,y))}{\sum_{x,y}\mathbbm{1}(M^s(x,y) \wedge C^s_i(x,y))}\;(i\in\{+,-\})
\label{eq:CAM prototype}
\end{equation}
\begin{equation}
  p^s=\frac{\sum_{x,y}X^s_1(x,y)\mathbbm{1}(M^s(x,y))}{\sum_{x,y}\mathbbm{1}(M^s(x,y))}
\label{eq:MAP}
\end{equation}
where $X^s_1(x,y)$ indexes the support feature vector at spatial locations $(x,y)$, and $\mathbbm{1}(\cdot)$ is an indicator function.

\begin{figure}[t]
  \centering
  \includegraphics[width=0.9\linewidth]{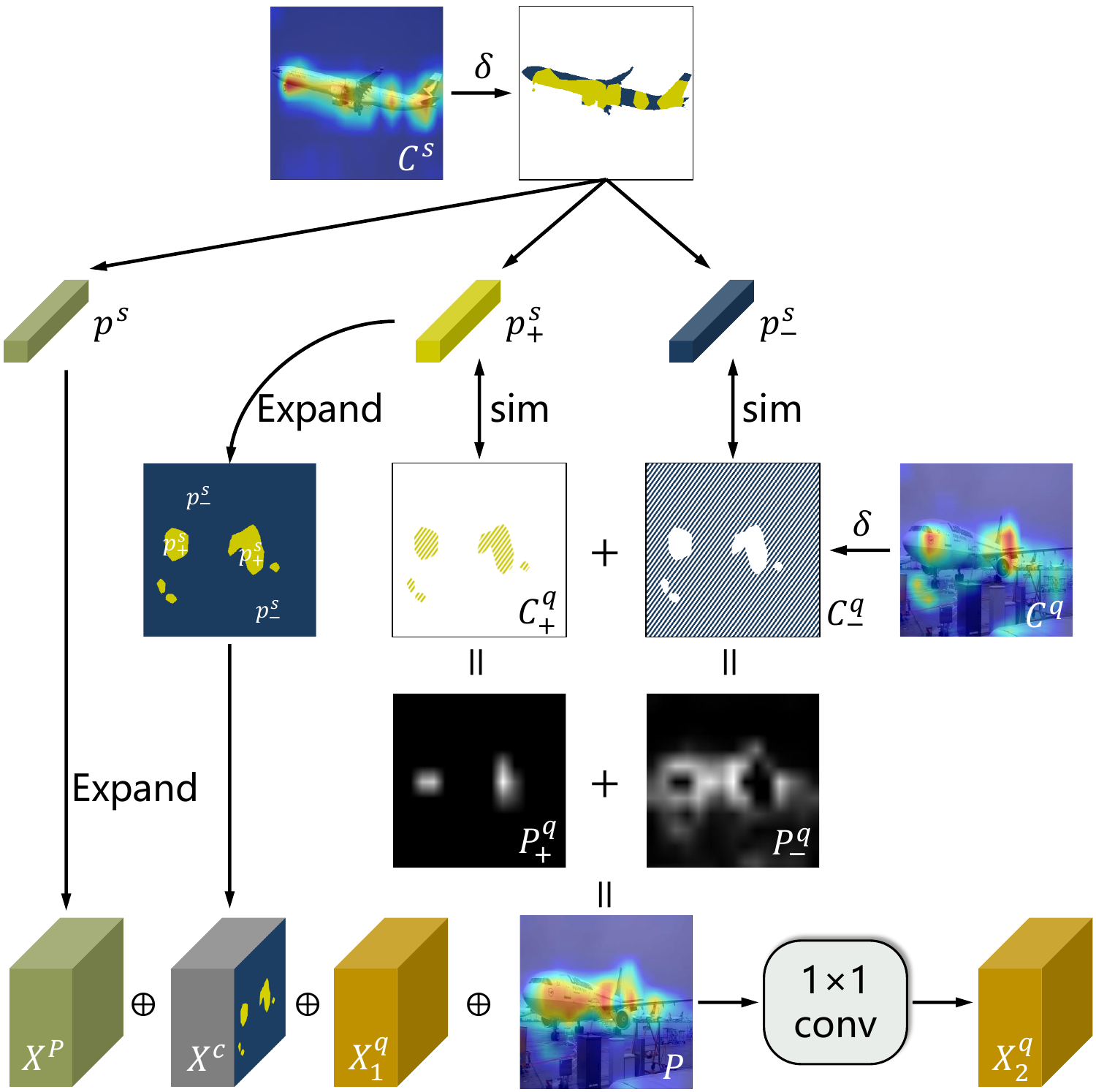}
  \caption{PMGM workflow: support/query CAM heatmaps are thresholded by $\delta$ into $C^s_+$, $C^s_-$, $C^q_+$, $C^q_-$; MAP on support foreground and its two sub-regions generates prototypes $p^s$, $p^s_+$, $p^s_-$; similarities between $p^s_+/C^q_+$ and $p^s_-/C^q_-$ yield prior map $P$; expanded $p^s$, expanded padded $p^s_+/p^s_-$, $X^q_1$, and $P$ are concatenated and fed into 1×1 convolution to obtain new $X^q_2$.}
  \label{fig:pmgm}
\end{figure}

\subsubsection{Prior Map Generation}
To generate the prior maps for the core discriminative region ($P^q_+$ for $C^q_+$) and local detail region ($P^q_-$ for $C^q_-$) in the query image, we build upon the baseline's single prior map generation pipeline and compute the feature similarity between query region-specific features and the support region prototypes ($p^s_+$ and $p^s_-$), respectively. Specifically, the foreground probability of each pixel in the query regions is quantified via cosine similarity and min-max normalization:
\begin{equation}
\begin{split}
  P^q_i(x,y)=\frac{(X^q_1(x,y))^Tq^s_i}{\Vert X^q(x,y)\Vert\Vert q^s_i\Vert}\;((x,y)\in C^q_i, i\in\{+,-\}),
\end{split}
\label{eq:cosine_cam}
\end{equation}
where $\Vert\cdot\Vert$ is the euclidean distance. $P^q_+(x,y)$ and $P^q_-(x,y)$ represent the foreground probability of pixel $(x,y)$ in the area $C^q_+$ and $C^q_-$, respectively.

Since the core discriminative region $C^q_+$ and local detail region $C^q_-$ of the query image are mutually complementary, we sum their respective prior maps to generate the full-image prior map. The resulting prior map $P$ is denoted as
\begin{equation}
P=P^q_++P^q_-.
\end{equation}

Building on the aforementioned prototypes and the full-image prior map $P$ derived above, the model fuses the core discriminative prototype $p^s_+$ and local detail prototype $p^s_-$ into the query feature space. Specifically, $p^s_+$ and $p^s_-$ are filled into their corresponding response areas $C^q_+$ and $C^q_-$, respectively:
\begin{equation}
  X^c=X^s_+\odot C^q_++X^s_-\odot C^q_-,
\end{equation}
where $X^s_+$ and $X^s_-$ denote the spatially expanded versions of $p^s_+$ and $p^s_-$, respectively.

Finally, the new feature $X^q_2$ of the query branch is generated by a 1×1 convolution $\mathcal{F}_{1\times1}$ that integrates the original feature, the foreground prototype, the core discriminative prototype, the local detail prototype, and the prior map. The entire pipeline is illustrated in Fig. \ref{fig:pmgm}. The support branch's new feature $X^s_2$ is generated in the same way.

\subsection{Directional Dropout Module (DDM)}
\label{subsec:Decoder}
The DDM is designed to address intra-image interference caused by directionally masking two incorrect support→query mapping: query interference (false positive) and support interference (false negative).
Our model incorporates support branch decoder to learn to refine the support prior map to support ground truth, guiding query branch decoder to segment query mask. At each decoder block with the index i, the inputs of both query and support branches are flattened as $X^q_t\in\mathbb{R}^{h_qw_q\times c}$ and $X^s_t\in\mathbb{R}^{h_sw_s\times c}$ respectively. Query features $X^q_t$ are denoted as \textit{Query} and support features $X^s_t$ are denoted as \textit{Key} and \textit{Value} in cross attention:
\begin{equation}
\begin{split}
  A_t=\frac{QK^\top}{\sqrt{c}},
  Q=X^q_tW^Q_t,
  K=X^s_tW^K_t,
  V=X^s_tW^V_t,
\end{split}
\end{equation}
where $W^Q_t, W^K_i,W^V_i\in\mathbb{R}^{c\times c}$ are the learnable weights to generate \textit{Query}, \textit{Key} and \textit{Value} in the $t$-th cross attention block.

The mispredicted regions in the query image can be categorized into two types of cases:
\begin{itemize}
\item False positive: the GT query mask of pixel $i$ corresponds to the background, whereas the most similar feature from the support pixel $j$ corresponds to the foreground (Fig. \ref{fig:model_arch} DDM, $(0,1)$);
\item False negative: the GT query mask $i$ corresponds to the foreground, whereas the most similar feature from the support features $j$ corresponds to the background (Fig. \ref{fig:model_arch} DDM, $(1,0)$).
\end{itemize}
Since the values in the cross attention map $A_t$ reflect the similarity between the corresponding query features and support features, based on the above mispredicted region attribution, DDM iterates through each query pixel $i$, retrieves the support pixel $j$ corresponding to the maximum value in row $A_t[i]$, checks their masks, and disrupts the inconsistent support-query mask pairs ($M^q_i\neq M^s_j$) to prevent the propagation of confusing information.

During the test phase, the DDM reverts to the traditional cross-attention mechanism due to the absence of the query GT mask.
To address the potential performance degradation caused by the training-test distribution mismatch, we propose a dynamic dropout strategy. DDM takes the mask frequency as the dynamic dropout rate $r$, and subsequently multiplies the remaining unmasked elements by $\frac{1}{1-r}$. This ensures consistent feature expectations between the training and testing phases, thereby guaranteeing distribution consistency across the two phases.

In short, DDM applies fine-grained pixel-wise masking, its overall pipeline is illustrated in Algorithm \ref{al:DDM}.

\begin{algorithm}
\caption{Directional Dropout Module}\label{al:DDM}
\renewcommand{\KwData}{\textbf{Input:}}
\renewcommand{\KwResult}{\textbf{Output:}}
\KwData{\\cross attention map $A\in\mathbb{R}^{h_qw_q\times h_sw_s}$, \\flatten support mask $M^s\in\mathbb{R}^{h_sw_s}$, \\flatten query mask $M^q\in\mathbb{R}^{h_qw_q}$}

\KwResult{masked cross attention map $A'\in\mathbb{R}^{h_qw_q\times h_sw_s}$}

Initialize cross attention mask $A^s\in\mathbb{R}^{h_qw_q\times h_sw_s}$ and mask ratio $r$ with 0\;
\For{$i=1\ to\ h_qw_q$}{
$j=\arg\max_{k \in \{1,2,...,h_sw_s\}} A[i,k]$\;
\If{$M^q[i]\neq M^s[j]$}{$A^s[i,j]=-\infty$\;$r=r+\frac{1}{h_qw_q\times h_sw_s}$\;}
}
$A'=A+A^s$\;
$A'=A'\times\frac{1}{1-r}$\;
\end{algorithm}

The query branch $(t+1)$-th block input $X^q_{t+1}$ is:
\begin{equation}
\begin{split}
    A'_t&=\text{DDM}(A_t, M^s, M^q), \\
    X^q_{t+1}&=\text{MLP}(\text{softmax}(A'_t)V).
\end{split}
\end{equation}

Finally, the overall loss can be computed using
\begin{equation}
\mathcal{L}=\text{BCE}(\hat{M}^q, M^q)+\frac{1}{N}\sum_{t=1}^N\text{BCE}(\hat{M}_t^q, M^q),
\end{equation}
where N denotes the number of cross blocks, and $\hat{M}^q$ as well as $\hat{M}_t^q$ represent the final prediction and intermediate prediction of the $t$-th decoder block, respectively.
\begin{table*}[htbp]
\centering
\caption{Quantitative comparison results on the PASCAL-5$^i$ dataset. The best and the second best mIoU (in \%) are highlighted with \textbf{bold} and \underline{underline}, respectively.}
\label{tab:pascal}
\resizebox{\linewidth}{!}{
\begin{tabular}{ccccccccccccc}
\hline
\multicolumn{1}{c|}{\multirow{2}{*}{Methods}} & \multicolumn{6}{c|}{1-shot} & \multicolumn{6}{c}{5-shot} \\ \cline{2-13} 
\multicolumn{1}{c|}{} & Fold-0 & Fold-1 & Fold-2 & \multicolumn{1}{c|}{Fold-3} & Mean & \multicolumn{1}{c|}{FB-IoU} & Fold-0 & Fold-1 & Fold-2 & \multicolumn{1}{c|}{Fold-3} & Mean & FB-IoU \\ \hline
\multicolumn{13}{c}{ResNet 50 Backbone} \\ \hline
\multicolumn{1}{c|}{CyCTR$_{2021}$\cite{CyCTR}} & 65.7 & 71.0 & 59.5 & \multicolumn{1}{c|}{59.7} & 64.0 & \multicolumn{1}{c|}{-} & 69.3 & 73.5 & 63.8 & \multicolumn{1}{c|}{63.5} & 67.5 & - \\
\multicolumn{1}{c|}{PFENet++$_{2024}$\cite{PFENet++}} & 63.3 & 71.0 & 65.9 & \multicolumn{1}{c|}{59.6} & 64.9 & \multicolumn{1}{c|}{76.8} & 66.1 & 75.0 & {\underline{74.1}} & \multicolumn{1}{c|}{64.3} & 69.9 & 81.1 \\
\multicolumn{1}{c|}{TBS$_{2024}$\cite{TBS}} & 68.5 & 72.0 & 63.8 & \multicolumn{1}{c|}{59.5} & 65.9 & \multicolumn{1}{c|}{77.7} & {72.3} & {74.1} & 68.4 & \multicolumn{1}{c|}{67.2} & {70.5} & 81.3 \\
\multicolumn{1}{c|}{CGMGM$_{2024}$\cite{CGMGM}} & 71.1 & 75.0 & {69.6} & \multicolumn{1}{c|}{63.7} & 69.9 & \multicolumn{1}{c|}{80.5} & 71.8 & \underline{78.9} & 69.1 & \multicolumn{1}{c|}{68.6} & 72.1 & 83.1 \\
\multicolumn{1}{c|}{ABCB$_{2024}$\cite{ABCB}} & \underline{72.9} & {\underline{76.0}} & 69.5 & \multicolumn{1}{c|}{{64.0}} & {70.6} & \multicolumn{1}{c|}{-} & \underline{74.4} & 78.0 & 73.9 & \multicolumn{1}{c|}{68.3} & 73.6 & - \\
\multicolumn{1}{c|}{AENet$_{2024}$\cite{AENet}} & 71.3 & 75.9 & 68.6 & \multicolumn{1}{c|}{65.4} & 70.3 & \multicolumn{1}{c|}{-} & {73.9} & {77.8} & 73.3 & \multicolumn{1}{c|}{{\underline{72.0}}} & {74.2} & - \\
\multicolumn{1}{c|}
{HMNet$_{2024}$\cite{HMNet}} & 72.2 & 75.4 & 70.0 & \multicolumn{1}{c|}{63.9} & 70.4 & \multicolumn{1}{c|}{\underline{81.6}} & {74.2} & {77.3} & \underline{74.1} & \multicolumn{1}{c|}{70.9} & {74.1} & \underline{84.4} \\
\multicolumn{1}{c|}{OCNet$_{2025}$\cite{OCNet}} & \textbf{73.5} & 75.9 & \underline{71.1} & \multicolumn{1}{c|}{\underline{64.9}} & \underline{71.4} & \multicolumn{1}{c|}{-} & {\textbf{75.9}} & {77.1} & \underline{74.1} & \multicolumn{1}{c|}{{70.9}} & {\underline{74.5}} & - \\ \hline
\multicolumn{1}{c|}{BAM$_{2023}$\cite{BAM}} & 69.2 & 74.7 & 67.8 & \multicolumn{1}{c|}{61.7} & 68.3 & \multicolumn{1}{c|}{-} & 71.8 & 75.7 & 72.0 & \multicolumn{1}{c|}{67.5} & 71.8 & - \\
\multicolumn{1}{c|}{IIR(Ours)} & 72.0 & \textbf{77.5} & \textbf{74.1} & \multicolumn{1}{c|}{\textbf{68.1}} & \textbf{72.9} & \multicolumn{1}{c|}{\textbf{83.3}} & 73.6 & \textbf{79.8} & \textbf{76.2} & \multicolumn{1}{c|}{\textbf{72.6}} & \textbf{75.6} & \textbf{85.1} \\ \hline
\multicolumn{13}{c}{ResNet 101 Backbone} \\ \hline
\multicolumn{1}{c|}{CyCTR$_{2021}$\cite{CyCTR}} & 67.2 & 71.1 & 57.6 & \multicolumn{1}{c|}{59.0} & 63.7 & \multicolumn{1}{c|}{-} & 71.0 & 75.0 & 58.5 & \multicolumn{1}{c|}{65.0} & 67.4 & - \\
\multicolumn{1}{c|}{DBMNet$_{2024}$\cite{DBMNet}} & 64.2 & 72.1 & 64.8 & \multicolumn{1}{c|}{58.9} & 65.0 & \multicolumn{1}{c|}{-} & 68.6 & 73.3 & 69.0 & \multicolumn{1}{c|}{64.2} & 68.8 & - \\
\multicolumn{1}{c|}{PFENet++$_{2024}$\cite{PFENet++}} & 63.1 & 72.4 & 63.4 & \multicolumn{1}{c|}{62.2} & 65.3 & \multicolumn{1}{c|}{75.5} & 67.2 & 76.1 & \textbf{75.5} & \multicolumn{1}{c|}{67.2} & 71.5 & 82.7 \\
\multicolumn{1}{c|}{ABCB$_{2024}$\cite{ABCB}} & \textbf{73.0} & {\underline{76.0}} & {\underline{69.7}} & \multicolumn{1}{c|}{{\underline{69.2}}} & {\underline{71.9}} & \multicolumn{1}{c|}{-} & \textbf{74.8} & {\underline{78.5}} & 73.6 & \multicolumn{1}{c|}{{\underline{72.6}}} & {\underline{74.8}} & - \\ \hline
\multicolumn{1}{c|}{BAM$_{2023}$\cite{BAM}} & 69.9 & 75.4 & 67.1 & \multicolumn{1}{c|}{62.1} & 68.6 & \multicolumn{1}{c|}{80.2} & 72.6 & 77.1 & 70.7 & \multicolumn{1}{c|}{69.8} & 72.5 & 84.1 \\
\multicolumn{1}{c|}{IIR(Ours)} & {\underline{72.8}} & \textbf{77.9} & \textbf{71.5} & \multicolumn{1}{c|}{\textbf{70.0}} & \textbf{73.1} & \multicolumn{1}{c|}{\textbf{82.1}} & {\underline{74.1}} & \textbf{79.6} & {\underline{75.3}} & \multicolumn{1}{c|}{\textbf{73.3}} & \textbf{75.6} & \textbf{84.7} \\ \hline
\end{tabular}}
\end{table*}%

\begin{table*}[]
\centering
\caption{Quantitative comparison results on the COCO-20$^i$ dataset. The best and the second best mIoU (in \%) are highlighted with \textbf{bold} and \underline{underline}, respectively.}
\label{tab:COCO}
\begin{tabular}{ccccccccccccc}
\hline
\multicolumn{1}{c|}{\multirow{2}{*}{Methods}} & \multicolumn{6}{c|}{1-shot} & \multicolumn{6}{c}{5-shot} \\ \cline{2-13}
\multicolumn{1}{c|}{} & Fold-0 & Fold-1 & Fold-2 & \multicolumn{1}{c|}{Fold-3} & Mean & \multicolumn{1}{c|}{FB-IoU} & Fold-0 & Fold-1 & Fold-2 & \multicolumn{1}{c|}{Fold-3} & Mean & FB-IoU \\ \hline
\multicolumn{13}{c}{ResNet 50 Backbone}
\\ \hline
\multicolumn{1}{c|}{CyCTR$_{2021}$\cite{CyCTR}} & 38.9 & 43.0 & 39.6 & \multicolumn{1}{c|}{39.8} & 40.3 & \multicolumn{1}{c|}{-} & 41.1 & 48.9 & 45.2 & \multicolumn{1}{c|}{47.0} & 45.6 & - \\
\multicolumn{1}{c|}{CGMGM$_{2024}$\cite{CGMGM}} & {\underline{47.1}} & 49.3 & 48.8 & \multicolumn{1}{c|}{44.4} & 47.4 & \multicolumn{1}{c|}{-} & 50.3 & 54.6 & 51.3 & \multicolumn{1}{c|}{51.8} & 52.0 & - \\
\multicolumn{1}{c|}{ABCB$_{2024}$\cite{ABCB}} & 44.2 & 54.0 & 52.1 & \multicolumn{1}{c|}{49.8} & 50.0 & \multicolumn{1}{c|}{-}& 50.5 & 59.1 & 57.0 & \multicolumn{1}{c|}{53.6} & 55.1 & - \\
\multicolumn{1}{c|}{HMNet$_{2024}$\cite{HMNet}} & 45.5 & \textbf{58.7} & \textbf{52.9} & \multicolumn{1}{c|}{\textbf{51.4}} & \underline{52.1} & \multicolumn{1}{c|}{\underline{74.5}} & \underline{53.4} & \underline{64.6} & \textbf{60.8} & \multicolumn{1}{c|}{\underline{56.8}} & \underline{58.9} & \textbf{77.6} \\
\multicolumn{1}{c|}{OCNet$_{2025}$\cite{OCNet}} & 45.9 & 56.9 & {\textbf{52.9}} & \multicolumn{1}{c|}{50.4} & 51.5 & \multicolumn{1}{c|}{73.7} & 52.7 & 63.1 & 57.4 & \multicolumn{1}{c|}{54.8} & 57.0 & 76.8 \\ \hline
\multicolumn{1}{c|}{BAM$_{2023}$\cite{BAM}} & 43.8 & 51.4 & 47.9 & \multicolumn{1}{c|}{44.5} & 46.9 & \multicolumn{1}{c|}{-} & 49.8 & 55.4 & 52.3 & \multicolumn{1}{c|}{50.2} & 51.9 & - \\
\multicolumn{1}{c|}{IIR(Ours)} & \textbf{49.7} & {\underline{58.1}} & 52.3 & \multicolumn{1}{c|}{\textbf{51.4}} & \textbf{53.1} & \multicolumn{1}{c|}{\textbf{75.4}} & \textbf{53.7} & {\textbf{65.2}} & \underline{59.6} & \multicolumn{1}{c|}{\textbf{57.3}} & \textbf{59.0} & \underline{77.3} \\ \hline
\multicolumn{13}{c}{ResNet 101 Backbone}  \\ \hline
\multicolumn{1}{c|}{HPA$_{2023}$\cite{HPA}} & 43.2 & 50.5 & 46.2  & \multicolumn{1}{c|}{46.3}  & 48.8 & \multicolumn{1}{c|}{-} & 49.4 & 58.4 & 52.5 & \multicolumn{1}{c|}{50.9} & 52.8 & - \\
\multicolumn{1}{c|}{DBMNet$_{2024}$\cite{DBMNet}} & 41.8 & 45.6 & 43.2  & \multicolumn{1}{c|}{41.3}  & 43.0 & \multicolumn{1}{c|}{-} & 46.7 & 50.5 & 48.8 & \multicolumn{1}{c|}{44.7} & 47.7 & - \\
\multicolumn{1}{c|}{ABCB$_{2024}$\cite{ABCB}} & {\underline{46.0}} & {\underline{56.3}} & {\underline{54.3}}  & \multicolumn{1}{c|}{{\underline{51.3}}}  & {\underline{51.5}} & \multicolumn{1}{c|}{-} & {\underline{51.6}} & {\underline{63.5}} & {\underline{62.8}} & \multicolumn{1}{c|}{{\underline{57.2}}} & {\underline{58.8}} & - \\ \hline
\multicolumn{1}{c|}{BAM$_{2023}$\cite{BAM}} & 45.2 & 55.1 & 48.7  & \multicolumn{1}{c|}{45.0} & 48.5 & \multicolumn{1}{c|}{-} & 48.3 & 58.4 & 52.7 & \multicolumn{1}{c|}{51.4} & 52.7 & - \\
\multicolumn{1}{c|}{IIR(Ours)} & \textbf{50.4} & \textbf{58.6} & \textbf{56.1}  & \multicolumn{1}{c|}{\textbf{54.0}} & \textbf{54.8} & \multicolumn{1}{c|}{74.8} & \textbf{52.8} & \textbf{64.7} & \textbf{63.0} & \multicolumn{1}{c|}{\textbf{57.4}} & \textbf{59.5} & 77.9 \\ \hline
\end{tabular}
\end{table*}

\begin{table}[htbp]
\centering
\caption{Quantitative comparison results on the LVIS-92$^i$ dataset in terms of mIoU (\%) under 1-shot and 5-shot settings.}
\label{tab:lvis}
\resizebox{\linewidth}{!}{
\begin{tabular}{cc|cc}
\hline
Method & Type & 1-shot & 5-shot \\ \hline
HSNet$_{2021}$\cite{HSNet} & correlation-based & 17.4 & 22.9 \\
VAT$_{2022}$\cite{VAT} & correlation-based & 18.5 & 22.7 \\
Painter$_{2023}$\cite{Painter} & VLM-based & 10.5 & 10.9 \\
SegGPT$_{2023}$\cite{SegGPT} & VLM-based & 18.6 & 25.4 \\
\hline
baseline & Prototype-based & 14.6 & 20.9 \\
IIR (Ours) & Prototype-based & 21.5 & 26.3 \\ \hline
\end{tabular}}
\end{table}

\section{Experiment}
\label{sec:Exp}
We evaluate the performance of our IIR model on 11 datasets across three tasks: \textbf{standard few shot segmentation (FSS)}, \textbf{object part few shot segmentation (OPFSS)}, and \textbf{cross domain few shot segmentation (CDFSS)}. These three tasks are respectively defined as pixel-wise segmentation of unseen novel categories based on a small number of annotated samples, fine-grained pixel-wise segmentation of specific parts of objects, and segmentation of novel categories in new domains.
Additionally, we conduct ablation experiments to validate the effectiveness of the two modules, PMGM and DDM. Experiments demonstrate that our model achieves excellent performance across different tasks and backbones.

\subsection{Datasets}
Datasets includes 9 existing FSS datasets across 3 tasks:
\begin{enumerate}
    \item FSS: PASCAL-5$^i$ \cite{SG-ONE}, COCO-20$^i$ \cite{SG-ONE}, LVIS-92$^i$ \cite{Matcher};
    \item OPFSS: PASCAL-Part \cite{Matcher}, PACO-Part \cite{Matcher};
    \item CDFSS: FSS-1000 \cite{FSS1000}, Deepglobe \cite{deepglobe}, ISIC2018 \cite{isic}, Chest X-Ray \cite{chest1,chest2}.
\end{enumerate}
More details are as shown in Appendix Section \ref{sec:dataset}.

Existing CDFSS comprises two medical datasets (ISIC-2018 for skin diseases and Chest X-Ray for chest diseases), failing to fully reflect CDFSS’s advantages. Their mask regions have clear boundaries, with annotation merely requiring boundary delineation, undermining the necessity of utilizing FSS (columns 2 and 3 of Fig. \ref{fig:cdfss}).
In contrast, fundus blood vessel segmentation faces small public datasets and high annotation difficulty: image contains dense small blood vessels, and mask annotation demands precise vascular trajectory delineation. Each image takes hours and substantial labor for repeated verification.
To better evaluate current CDFSS methods on complex topological and real-world medical tasks, we build two fundus blood vessel CDFSS benchmarks (Fives, Stare) based on public datasets; see Appendix Section \ref{sec:dataset} for details.

\subsection{Implementation Details}
For PASCAL-5$^i$ and COCO-20$^i$, our method adopts BAM \cite{BAM} as its baseline. For LVIS-92$^i$, PASCAL-part and PACO-Part, we follow the setting of Matcher \cite{Matcher}. For CDFSS datasets, our model takes IFA \cite {IFA} as its baseline. More details are shown in Appendix section \ref{sec:Implementation}.
We adopted mean intersection over union (mIoU) and foreground-background intersection over union (FB-IoU) as evaluation metrics.
All tests of the proposed model on all datasets are conducted using the PyTorch and run on an NVIDIA RTX 4090 GPU.

\subsection{Comparison With State-of-the-Arts}
\subsubsection{FSS}
In Table \ref{tab:pascal}, Table \ref{tab:COCO} and Table \ref{tab:lvis}, we compare our method with other SOTA approaches on PASCAL-5$^i$, COCO-20$^i$ and LVIS-92$^i$, respectively.

For PASCAL-5$^i$, our IIR achieves 72.9\% mIoU with ResNet-50 backbone and 73.1\% mIoU with ResNet-101 backbone for 1-shot segmentation, representing improvements of 4.6\% and 4.5\% in mIoU over the baseline, respectively. It outperforms previous SOTA results by 1.5\% and 1.2\%, respectively. 
For 5-shot segmentation, our method attains mIoU gains of 3.8\% and 3.1\% compared to the baseline when using ResNet-50 and ResNet-101 backbones, respectively, while surpassing SOTA methods by 1.1\% and 0.8\% in mIoU.

For COCO-20$^i$ results in Table \ref{tab:COCO}, our method achieves 53.1\% mIoU with ResNet-50 backbone and 54.8\% mIoU with ResNet-101 backbone for 1-shot segmentation, representing improvements of 6.2\% and 6.3\% in mIoU over the baseline, respectively. For 5-shot segmentation, our IIR attains mIoU gains of 7.1\% and 6.8\% compared to the baseline when using ResNet-50 and ResNet-101 backbones, respectively, while surpassing SOTA methods by 0.1\% and 0.7\% in mIoU.

For LVIS-92$^i$ results in Table \ref{tab:lvis}, compared with the baseline, our model achieves improvements of 6.9\% and 5.4\% under the 1-shot and 5-shot settings, respectively. In comparison with correlation-based methods, we obtain gains of 3.0\% and 3.4\% under the 1-shot and 5-shot scenarios, respectively. 
Additionally, we list various SAM-based methods, among which Matcher \cite{Matcher} exhibit stronger performance. This may be attributed to the fact that the SAM-1B dataset \cite{SAM} contains a diverse range of image-mask pairs, thereby endowing methods based on it with robust FSS capabilities.

Based on the performance across these FSS datasets, we observe that 5-shot settings achieve slightly inferior performance compared to 1-shot settings under all configurations. We attribute this phenomenon to the fact that 5-shot settings integrate five prior maps, resulting in higher robustness of the prior map than that in 1-shot settings. As the PMGM is designed to enhance the accuracy and robustness of the prior map, the performance improvement under 5-shot settings is relatively limited.

\begin{table*}[htbp]
\centering
\caption{Quantitative comparison results on the PASCAL-Part and PACO-Part dataset in terms of mIoU (\%).}
\label{tab:part fss}
\resizebox{\textwidth}{!}{%
\begin{tabular}{cc|ccccc|ccccc}
\hline
\multirow{2}{*}{Method} & \multirow{2}{*}{Type} & \multicolumn{5}{c|}{PASCAL-Part} & \multicolumn{5}{c}{PACO-Part} \\ \cline{3-12} 
 & & animal & indoor & person & \multicolumn{1}{c|}{vehicles} & mean & Fold-0 & Fold-1 & Fold-2 & \multicolumn{1}{c|}{Fold-3} & mean \\ \hline
HSNet$_{2021}$ \cite{HSNet} & correlation-based     & 21.2   & 53.0 & 20.2   & \multicolumn{1}{c|}{35.1}     & 32.4 & 22.0 & 22.9   & 26.0 & \multicolumn{1}{c|}{23.1}   & 23.5 \\
VAT$_{2022}$ \cite{VAT} & correlation-based     & 21.5   & 55.9   & 20.7   & \multicolumn{1}{c|}{36.1} & 33.6 & 13.7   & 12.5   & 15.0 & \multicolumn{1}{c|}{15.1}   & 14.1 \\
Painter$_{2023}$ \cite{Painter} & VLM-based & 20.2   & 49.5   & 17.6   & \multicolumn{1}{c|}{34.4} & 30.4 & 13.9   & 12.6   & 14.8   & \multicolumn{1}{c|}{12.7}   & 13.5 \\
SegGPT$_{2023}$ \cite{SegGPT} & VLM-based & 22.8   & 50.9   & 31.3   & \multicolumn{1}{c|}{38.0} & 35.8 & 19.4   & 20.5   & 23.8   & \multicolumn{1}{c|}{21.2}   & 21.2 \\
\hline
baseline & Prototype-based & 18.8   & 50.6   & 21.1   & \multicolumn{1}{c|}{29.8} & 30.1 & 19.8   & 20.8   & 24.4   & \multicolumn{1}{c|}{23.1}   & 22.0   \\
IIR (Ours) & Prototype-based & 24.5   & 54.2   & 25.8   & \multicolumn{1}{c|}{32.1} & 34.2 & 21.7 & 22.3   & 25.7   & \multicolumn{1}{c|}{23.9}   & 23.4 \\ \hline
\end{tabular}%
}
\end{table*}

\begin{figure*}[htbp]
  \centering
  \includegraphics[width=0.95\linewidth]{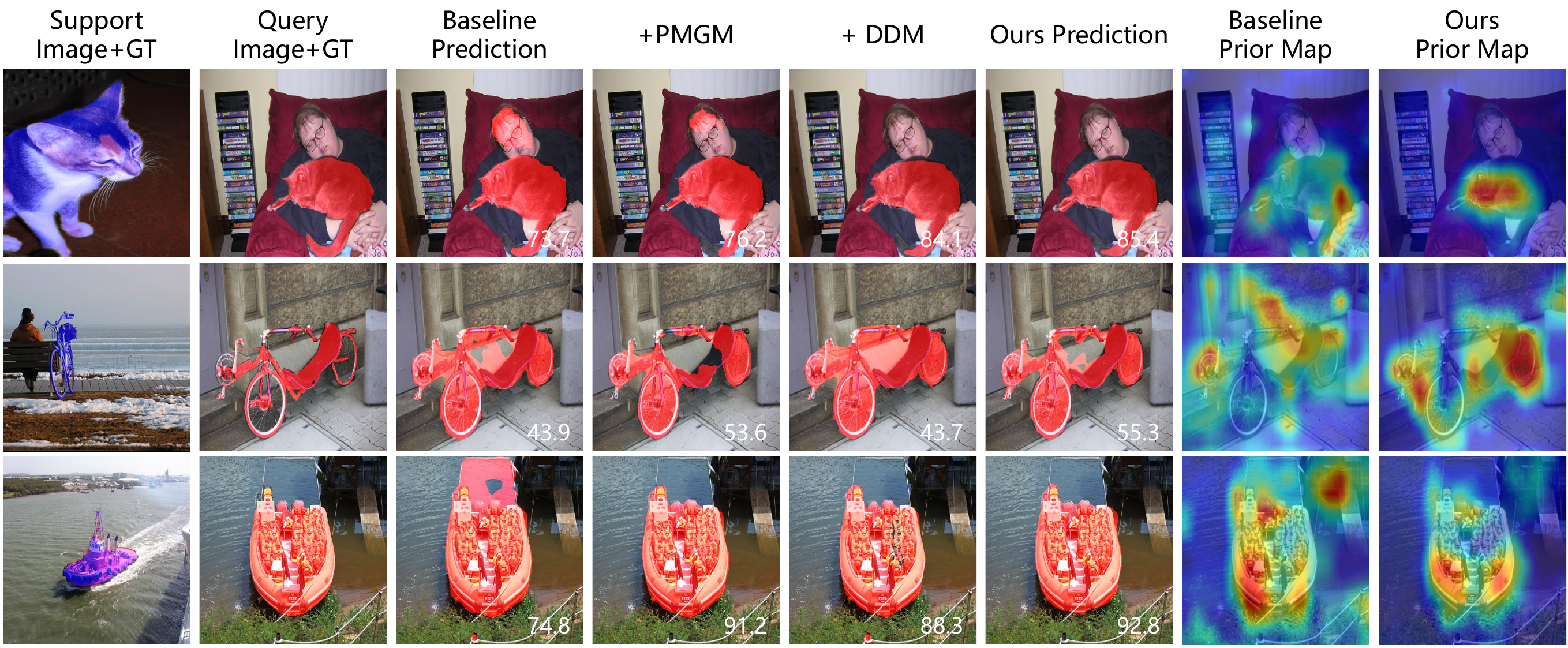}
  \caption{Visualization on the PASCAL-5$^i$ dataset. The first row is segmentation in similar backgrounds, the second row addresses segmentation within small foregrounds and the third row focuses on segmentation in salient foregrounds. The third column is the prediction of our baseline model BAM \cite{BAM}. The last two columns show our PMGM generates more accurate foreground cues. More visualization results are presented in the supplementary materials.}
  \label{fig:vis pascal}
\end{figure*}

\subsubsection{OPFSS}
In Table \ref{tab:part fss}, we present additional quantitative results on the PASCAL-Part and PACO-Part datasets. Among these results, our proposed IIR model outperforms the baseline method by a significant margin in all folds, providing a robust baseline result for future research. Specifically, our IIR surpasses the baseline by 4.1\% on PASCAL-Part and by 1.4\% on PACO-Part.
Furthermore, our model outperforms correlation-based methods in most cases, and even exceeds Painter and SegGPT.

\subsubsection{CDFSS}
Table \ref{tab:cdfss} presents the quantitative results of our IIR model for cross domain few shot segmentation under the 1-shot and 5-shot settings. It can be observed from the results that our model achieves significant performance improvements on most datasets. Compared with the baseline IFA:
on the ISIC dataset, the improvements are 2.6\% and 13.4\% under the 1-shot and 5-shot settings, respectively;
on the Chest X-ray dataset, the improvements are 16.6\% and 17.4\% under the 1-shot and 5-shot settings, respectively;
on the fives dataset, the improvements are 1.7\% and 1.5\% under the 1-shot and 5-shot settings, respectively;
on the stare dataset, the improvements are 4.0\% and 4.8\% under the 1-shot and 5-shot settings, respectively;
on average, the improvements over the baseline are 3.8\% and 6.1\% under the 1-shot and 5-shot settings, respectively.
However, our IIR method exhibits a significant performance drop on the Deepglobe dataset—particularly a 4.4\% decrease under the 1-shot setting. 
Through visualization analysis, we attribute this phenomenon to the significant domain gap between the training and test data: our model is trained on natural images, while Deepglobe is a remote sensing image dataset. The visual features and semantic patterns for prior map generation differ drastically between natural and remote sensing domains. This discrepancy weakens CAM’s ability to capture effective discriminative features for remote sensing targets, leading to reduced quality of the prior map and ultimately causing the performance drop on Deepglobe.
Furthermore, all CDFSS methods achieve relatively low performance on the fundus vessel segmentation dataset, which indicates that existing CDFSS approaches fail to accomplish the segmentation of complex topological structures. In this regard, FSS algorithms still require further improvement.

\begin{table*}[t]
\centering
\caption{Quantitative comparison results on the cross domain few shot segmentation datasets. the best and the second best miou (in \%) are highlighted with \textbf{bold} and \textit{underline}, respectively.
}
\label{tab:cdfss}
\resizebox{\linewidth}{!}{
\begin{tabular}{ccccccccccccccc}
\hline
\multicolumn{15}{c}{Source Domain: Pascal VOC 2012 → Target Domain: Below} \\ \hline
\multicolumn{1}{c|}{\multirow{2}{*}{method}} & \multicolumn{2}{c|}{Deepglobe} & \multicolumn{2}{c|}{ISIC} & \multicolumn{2}{c|}{Chest X-Ray} & \multicolumn{2}{c|}{FSS-1000} & \multicolumn{2}{c|}{Fives} & \multicolumn{2}{c|}{Stare} & \multicolumn{2}{c}{Average} \\ \cline{2-15} 
\multicolumn{1}{c|}{} & 1-shot & \multicolumn{1}{c|}{5-shot} & 1-shot & \multicolumn{1}{c|}{5-shot} & 1-shot & \multicolumn{1}{c|}{5-shot} & 1-shot & \multicolumn{1}{c|}{5-shot} & 1-shot & \multicolumn{1}{c|}{5-shot} & 1-shot & \multicolumn{1}{c|}{5-shot} & 1-shot & 5-shot \\ \hline
\multicolumn{1}{c|}{PATNet$_{2022}$\cite{PATNet}} & 37.9 & \multicolumn{1}{c|}{43.0} & 41.2 & \multicolumn{1}{c|}{53.6} & 66.6 & \multicolumn{1}{c|}{70.2} & 78.6 & \multicolumn{1}{c|}{81.2} & 11.7 & \multicolumn{1}{c|}{11.9} & 10.9 & \multicolumn{1}{c|}{10.9} & 41.2 & 45.1 \\
\multicolumn{1}{c|}{ABCDFSS$_{2024}$\cite{ABCDFSS}} & 42.6 & \multicolumn{1}{c|}{49.0} & 45.7 & \multicolumn{1}{c|}{53.3} & 79.8 & \multicolumn{1}{c|}{81.4} & 74.6 & \multicolumn{1}{c|}{76.2} & 11.3 & \multicolumn{1}{c|}{12.5} & 12.4 & \multicolumn{1}{c|}{12.8} & 44.4 & 47.5 \\
\multicolumn{1}{c|}{APM$_{2024}$\cite{APM}} & 40.9 & \multicolumn{1}{c|}{44.9} & 41.7 & \multicolumn{1}{c|}{51.2} & \underline{78.3} & \multicolumn{1}{c|}{\underline{82.8}} & 79.3 & \multicolumn{1}{c|}{\underline{81.8}} & 10.0 & \multicolumn{1}{c|}{10.1} & 10.4 & \multicolumn{1}{c|}{10.5} & 43.4 & 46.9 \\
\hline
\multicolumn{1}{c|}{IFA$_{2024}$\cite{IFA}} & \textbf{50.6} & \multicolumn{1}{c|}{\textbf{58.8}} & \underline{66.3} & \multicolumn{1}{c|}{\underline{69.8}} & 74.0 & \multicolumn{1}{c|}{74.6} & \underline{80.1} & \multicolumn{1}{c|}{\textbf{82.4}} & {\underline{20.6}} & \multicolumn{1}{c|}{{\underline{21.0}}} & {\underline{14.0}} & \multicolumn{1}{c|}{{\underline{14.0}}} & {\underline{50.9}} & {\underline{53.4}} \\
\multicolumn{1}{c|}{IIR (Ours)} & {\underline{46.2}} & \multicolumn{1}{c|}{{\underline{58.6}}} & \textbf{68.9} & \multicolumn{1}{c|}{\textbf{83.2}} & \textbf{90.6} & \multicolumn{1}{c|}{\textbf{92.0}} & \textbf{81.8} & \multicolumn{1}{c|}{81.6} & \textbf{22.3} & \multicolumn{1}{c|}{\textbf{22.5}} & \textbf{18.0} & \multicolumn{1}{c|}{\textbf{18.8}} & \textbf{54.7} & \textbf{59.5} \\ \hline
\end{tabular}}
\end{table*}

\begin{figure*}[htbp]
  \centering
  \includegraphics[width=0.95\textwidth]{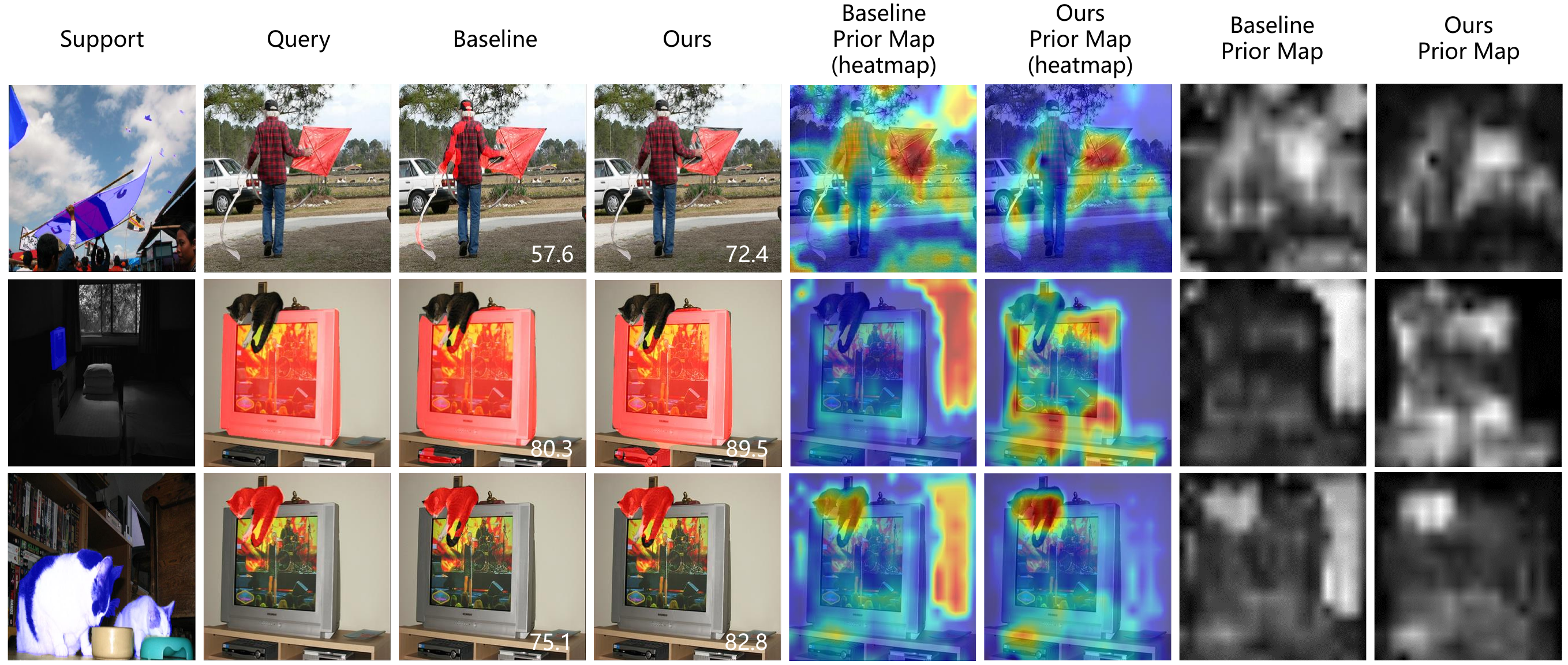}
  \caption{Visualization on COCO-20$^i$ dataset. Last two columns is the prior map on grayscale images. Our prior map outperforms baseline in foreground focusing and noise elimination. Last two rows have same query’s foreground distinction and we accurately targets corresponding foregrounds for better segmentation.}
  \label{fig:vis coco}
\end{figure*}

\begin{figure}[t]
  \centering
  \includegraphics[width=0.95\linewidth]{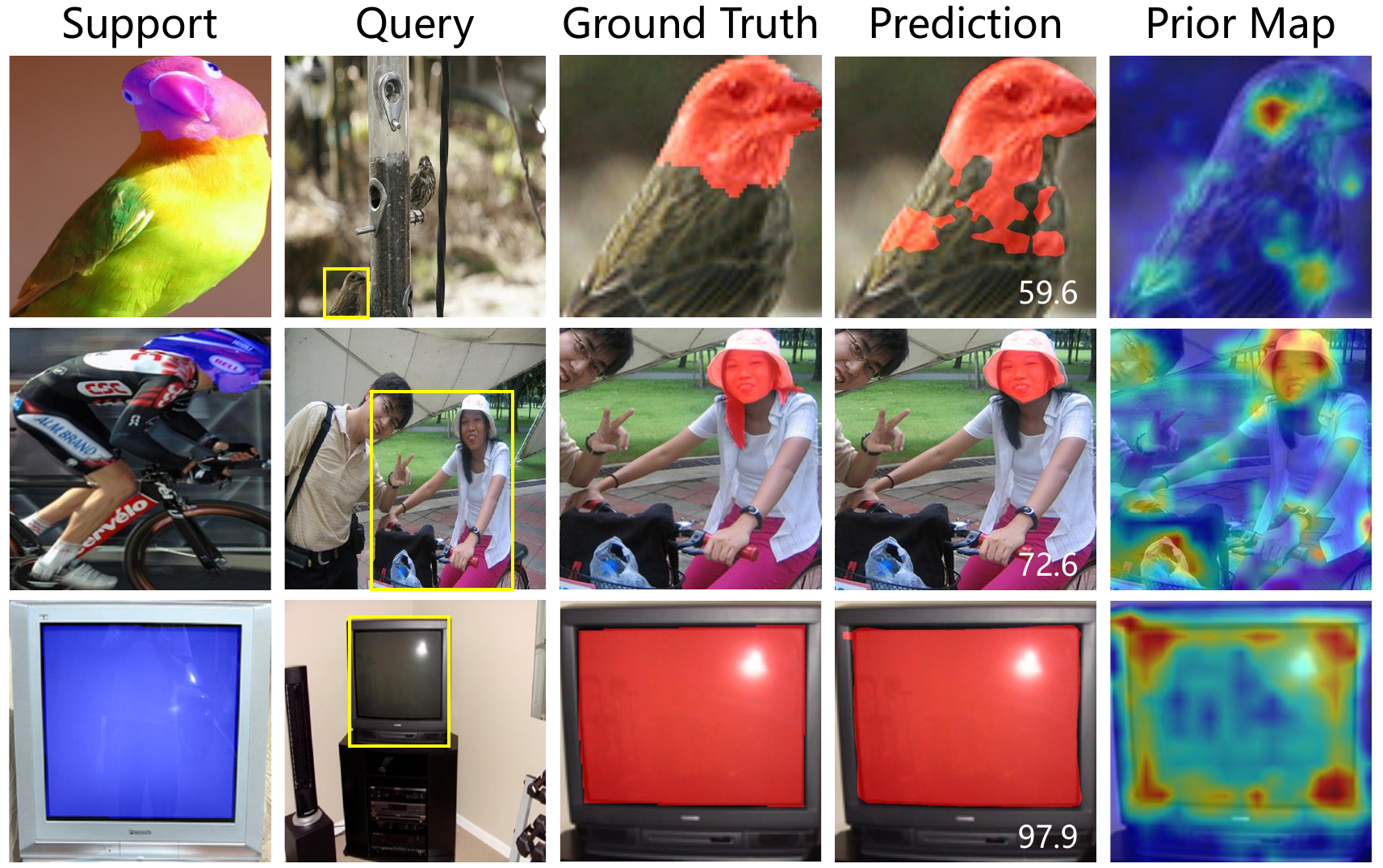}
  \caption{Visualization on PASCAL-Part dataset. The yellow box in the query image (second column) serves as model input (third column).}
  \label{fig:pascal-part}
\end{figure}

\begin{figure}[t]
  \centering
  \includegraphics[width=0.95\linewidth]{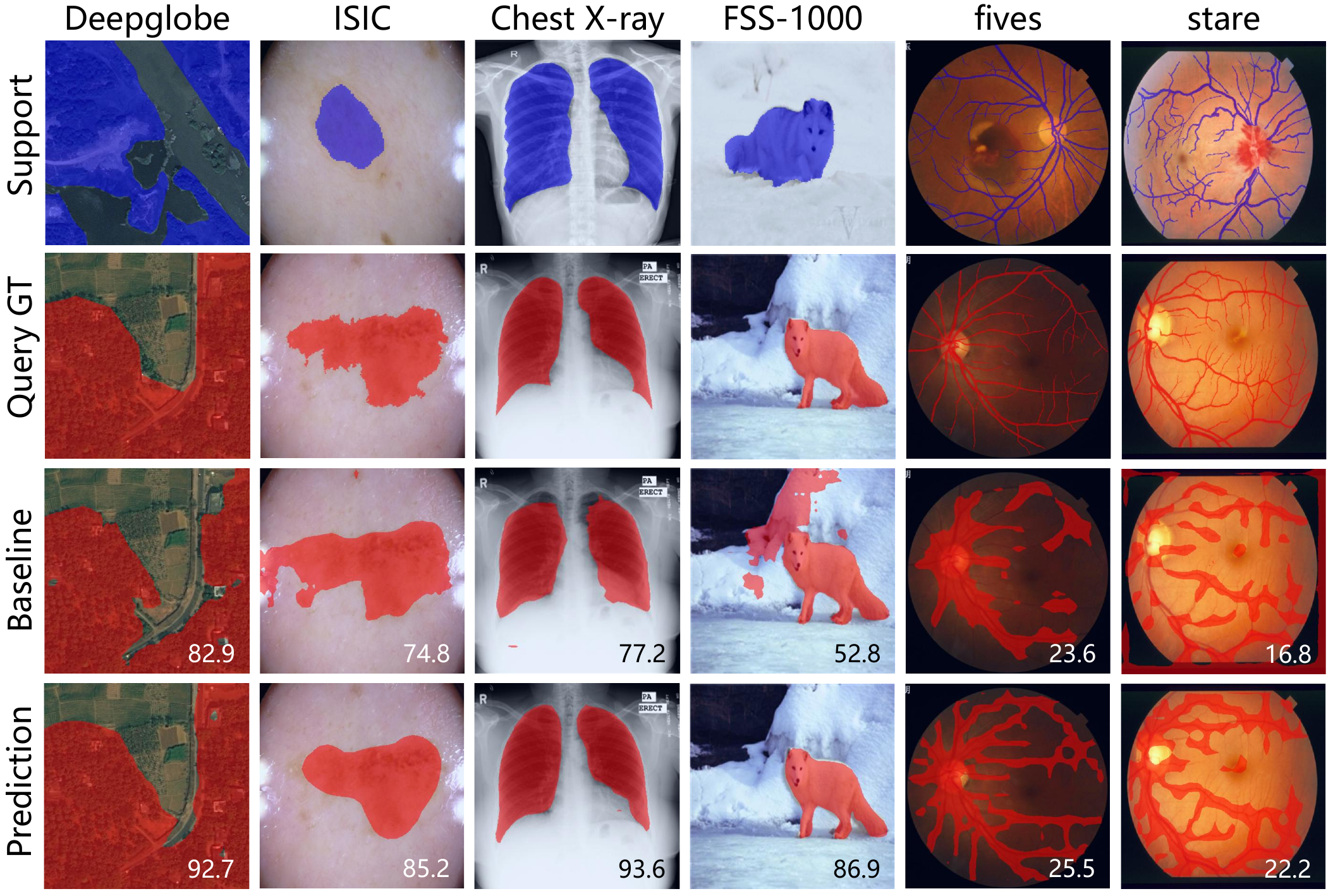}
  \caption{Visualization on cross domain datasets, including remote sensing domain, medical domain and natural domain.}
  \label{fig:cdfss}
\end{figure}

\subsection{Qualitative Results}
To more intuitively demonstrate the effectiveness of our algorithm, Fig. \ref{fig:vis pascal} presents the final prediction performance and prior map prompting capability of a single support-query image pair on the PASCAL-5$^i$ dataset under the ResNet50 1-shot setting, from the baseline to the IIR model with the incremental addition of each module. 
As illustrated in Fig. \ref{fig:vis pascal}, our model consistently delivers more accurate segmentation outcomes compared with the baseline across diverse challenging scenarios including query images with mixed foreground-background (row 1), complex topological structures (row 2) and significant dissimilarity between support and query foregrounds (row 3), by effectively concentrating the prior map on the query’s foreground. Further demonstrations are provided in Appendix Fig. \ref{fig:vis more pascal}.

We visualize the performance of our model on the COCO-20$^i$ dataset in Fig. \ref{fig:vis coco}. The last two columns present the effect of the prior map on grayscale images, from which it can be clearly observed that our prior map exhibits stronger foreground focusing ability and noise elimination ability compared with the baseline prior map. Additionally, in the last two rows, we compare the ability to distinguish different foregrounds for the same query image; the results show that our prior map can well focus on the corresponding foreground targets and ultimately achieve better segmentation results. 

We also conducted corresponding visualizations on the LVIS-92$^i$ dataset, as shown in Appendix Fig. \ref{fig:lvis}. The visualization results demonstrate that our method exhibits excellent segmentation performance for multiple targets, size-varying targets, and irregularly shaped targets. 
Fig. \ref{fig:pascal-part} and Appendix Fig. \ref{fig:paco-part} presents several representative segmentation results to illustrate typical scenarios in the object part few shot segmentation task, where the model is required to perform segmentation on cropped images as shown in the second column of figures.
The visualization results across cross domain datasets are shown in Fig. \ref{fig:cdfss}.

\begin{table*}[t]
\centering
\caption{BCE between the prior map and the ground truth. Lower BCE indicates greater similarity between the prior map and the ground truth. The subscript index is the standard deviation across ten random validation seed.}
\begin{tabular}{c|ccccc}
\hline
 & Fold-0 & Fold-1 & Fold-2 & Fold-3 & Mean  \\ \hline
Baseline & $0.2875_{0.0021}$ & $0.3550_{0.0149}$ & $0.3908_{0.0335}$ & $0.3698_{0.0284}$ & $0.3508_{0.0089}$ \\
IIR & $0.2457_{0.0018}$ & $0.2848_{0.0023}$ & $0.3029_{0.0023}$ & $0.2902_{0.0018}$ & $0.2809_{0.0007}$ \\ \hline
\end{tabular}
\label{tab:CE}
\end{table*}

\begin{table}[t]
\centering
\caption{Ablation Studies. `col -$\inf$' means masking the whole column to -$\inf$ when meeting inconsistency.}
\begin{tabular}{c|c|c|cc|c}
\hline
\multirow{2}{*}{PMGM} & \multirow{2}{*}{supp branch} & \multirow{2}{*}{CyCTR} & \multicolumn{2}{c|}{DDM} & \multirow{2}{*}{mIoU} \\
 & & & col -$\inf$ & cell -$\inf$ & \\ \hline
 & & & & & 70.5 \\
 & \checkmark & & & \checkmark & 71.4   \\
\checkmark & & & & & 71.3\\
\checkmark & \checkmark & & & & 71.6    \\
\checkmark & \checkmark & \checkmark & & & 71.8 \\
\checkmark & \checkmark & & \checkmark & & 71.8 \\
\checkmark & \checkmark & & & \checkmark & \textbf{72.9}\\ \hline
\end{tabular}
\label{table:ablation}
\end{table}

\subsection{Ablation Study}
We conduct ablation studies on the PASCAL-5$^i$ dataset with the ResNet50 backbone to evaluate the effectiveness of PMGM and DDM.
Table \ref{table:ablation} reveals incremental impact through row-wise comparisons. The baseline (BAM with equivalent decoders) presents mIoU results of 70.5\% (row 1). The mIoU improves 2.2\% by merely adding decoder number, as shown in Fig. \ref{fig:Saturation study}, demonstrating our architecture's inherent performance enhancement capabilities.
According to the third row of Table \ref{table:ablation}, the PMGM module, when used alone, improves the performance to 71.3\%. The DDM module, when used alone (as shown in the second row), enhances the performance by 0.9\%.

To further investigate and compare the performance details of the DDM and similar models, we first test the performance when only the support branch is introduced, which leads to a 0.3\% improvement (row 4 versus row 3). This result indicates that introducing support information can convey certain effective information to aid in segmentation.
Secondly, we introduce CyCTR and DDM for comparison. Compared with the baseline (fourth row), the introduction of CyCTR improves the performance by 0.2\% (fifth row).
The last row and the fifth row of Table \ref{table:ablation}, together with Table \ref{tab:Mask ratio}, present a performance comparison between our DDM and CyCTR. Our DDM can reduce the number of support-query inconsistent pairs, thereby achieving a 1.1\% improvement in mIoU. The reason lies in that our DDM can directionally mask only inconsistent support-query pairs, avoiding excessive column-wise masking.
We also conducted comparative tests on the internal parameter settings of DDM. Specifically, we compared the effects of the dropout strategy and the train-test consistent strategy of CyCTR, as shown in the 5-th and 6-th rows of Table \ref{table:ablation}. The results indicate that the dropout strategy can effectively align the gap between training and testing, and there is little difference in performance between this strategy and CyCTR’s train-test consistent strategy.
By comparing the 6th row with the last row, it can be seen that our masking strategy significantly reduces the mask ratio of the cross attention module (by $1 - \frac{1}{900} = 99.89\%$), thereby avoiding the drawback of information loss caused by excessive masking in CyCTR and improving the mIoU by 1.1\%.

Additional ablation experiments regarding the Decoder number, CAM method, and CAM threshold are provided in Appendix Section \ref{subsec:more ablation}.
\begin{figure}[t]
  \centering
  \includegraphics[width=\linewidth]{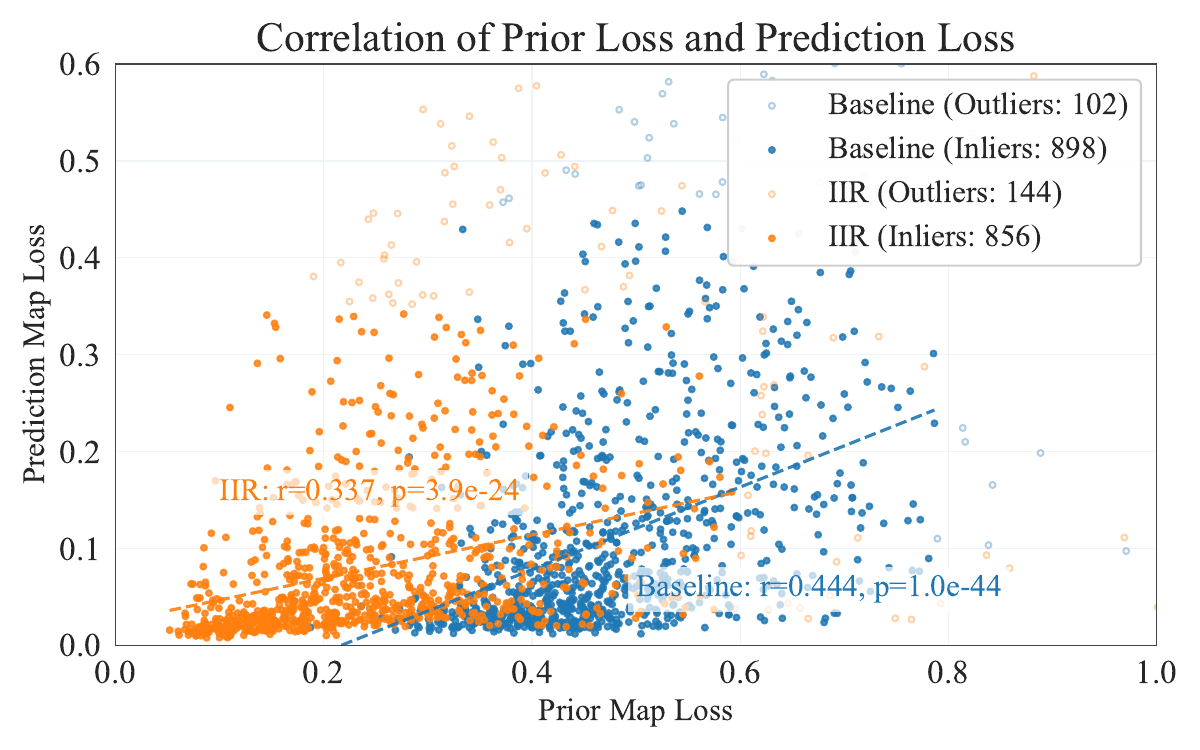}
  \caption{The BCE loss comparison on the PASCAL-5$^i$ datasets. Blue and orange points denote the baseline and our model’s samples, respectively. The results indicate that the BCE loss of the prior map and that of the prediction map exhibit a correlation and our model is far more accurate than the Baseline model, thereby improving the model performance.}
  \label{fig:bam_loss}
\end{figure}

\section{Discussion}
\label{sec:discussion}
\subsection{Prior Map}
\label{sec:motivation PM}
\textbf{Inter-Image Gap} with the use of a single prototype causes the prior map to be scattered and inaccurate. To quantify this issue, we calculated two sets of Binary Cross Entropy (BCE) losses for the baseline model \cite{BAM}: one between the prior map and ground truth (GT) results, and the other between the final prediction map and GT results. 
The Pearson Correlation Coefficient $r$ is used to quantify the direction and strength of the linear correlation between two variables.
The correlation coefficient $r$ in our experiment is 0.444, confirming that \textit{the lower the loss of the prior map, the lower the loss of the final prediction map}, as illustrated in Fig. \ref{fig:bam_loss}. 

Additionally, we performed quantitative analysis. Table \ref{tab:CE} demonstrates that the BCE loss of the prior map for IIR decreases by 19.93\% compared with the baseline, with a smaller standard deviation (the BCE loss of IIR is 0.2809±0.0007, while that of the baseline is 0.3508±0.00089). This indicates that the dual prototypes of ``foreground core features + local specific features”, generated by PMGM via CAM, effectively mitigate the background noise and foreground dispersion issues of the prior map. 
The prior map of IIR is more focused on foreground regions, thereby improves the model performance.

% 正文并排显示2*2张图
\begin{figure}[t]
  \centering
  % 第一行第一列子图
  \subfigure[The inconsistency ratio of CAB 1]{
    \includegraphics[width=0.225\textwidth]{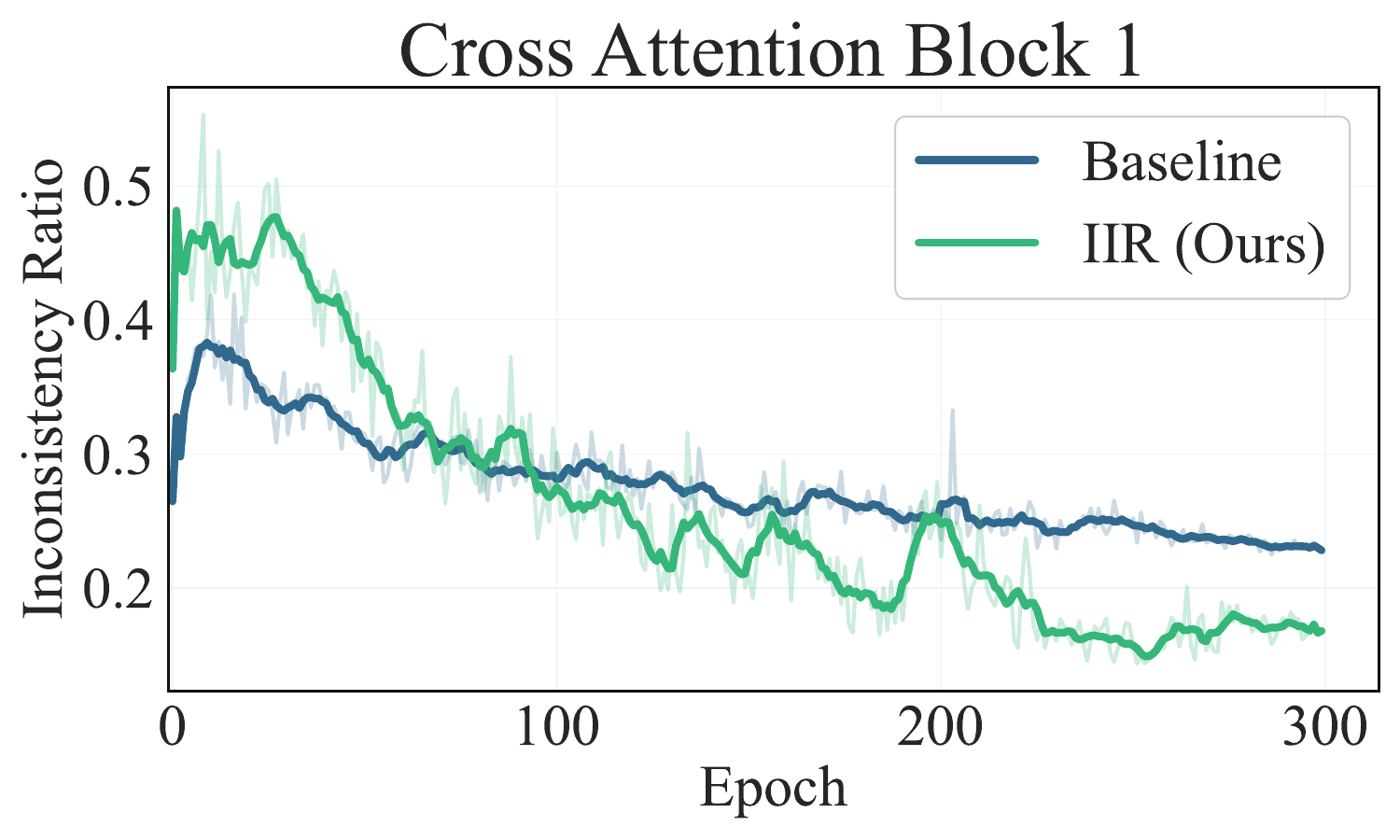}
    \label{subfig:CAB1}
  }
  \hfill % 第一行两子图之间的间距
  % 第一行第二列子图
  \subfigure[The inconsistency ratio of CAB 2]{
    \includegraphics[width=0.225\textwidth]{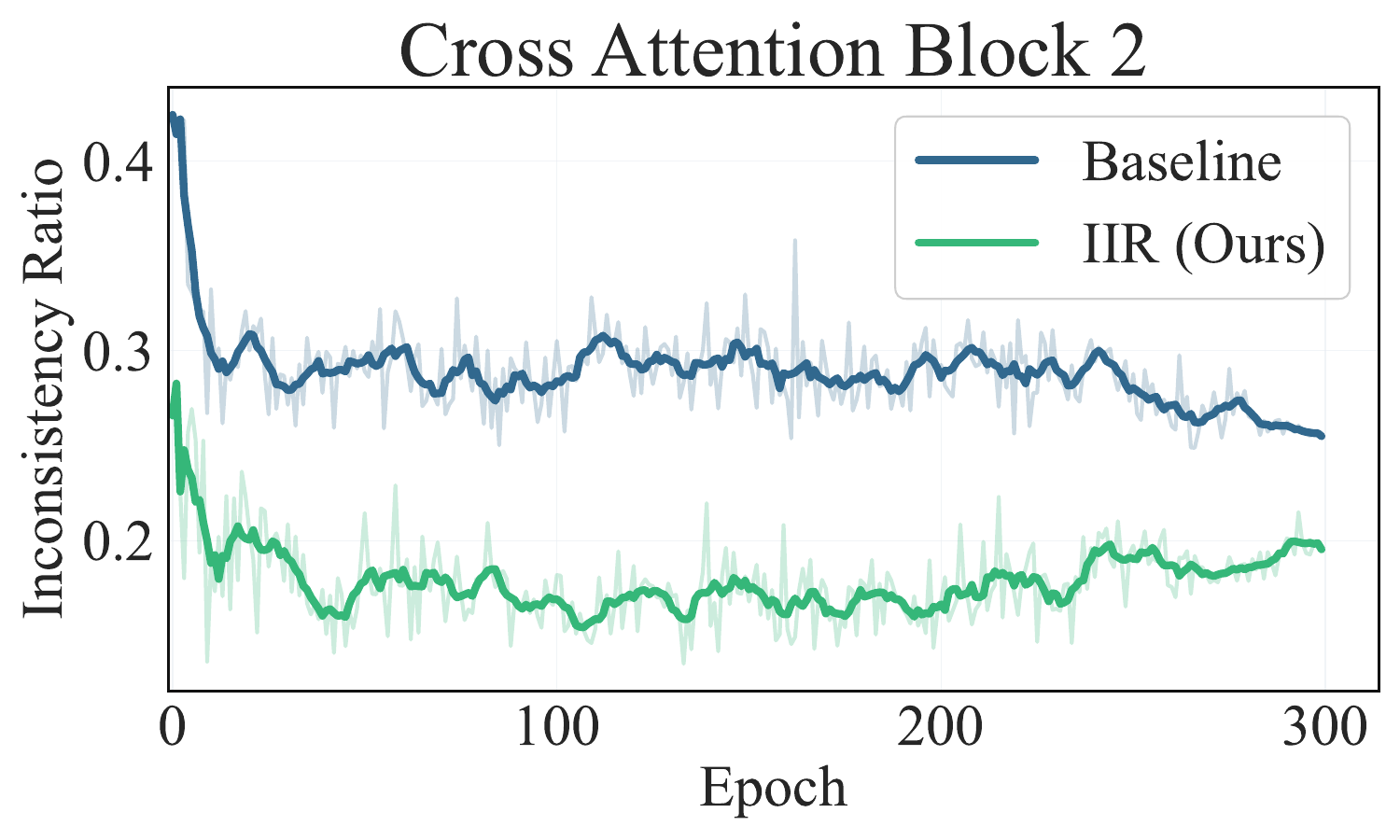}
    \label{subfig:CAB2}
  }
  \\
  % 第二行第一列子图
  \subfigure[The inconsistency ratio of CAB 3]{
    \includegraphics[width=0.225\textwidth]{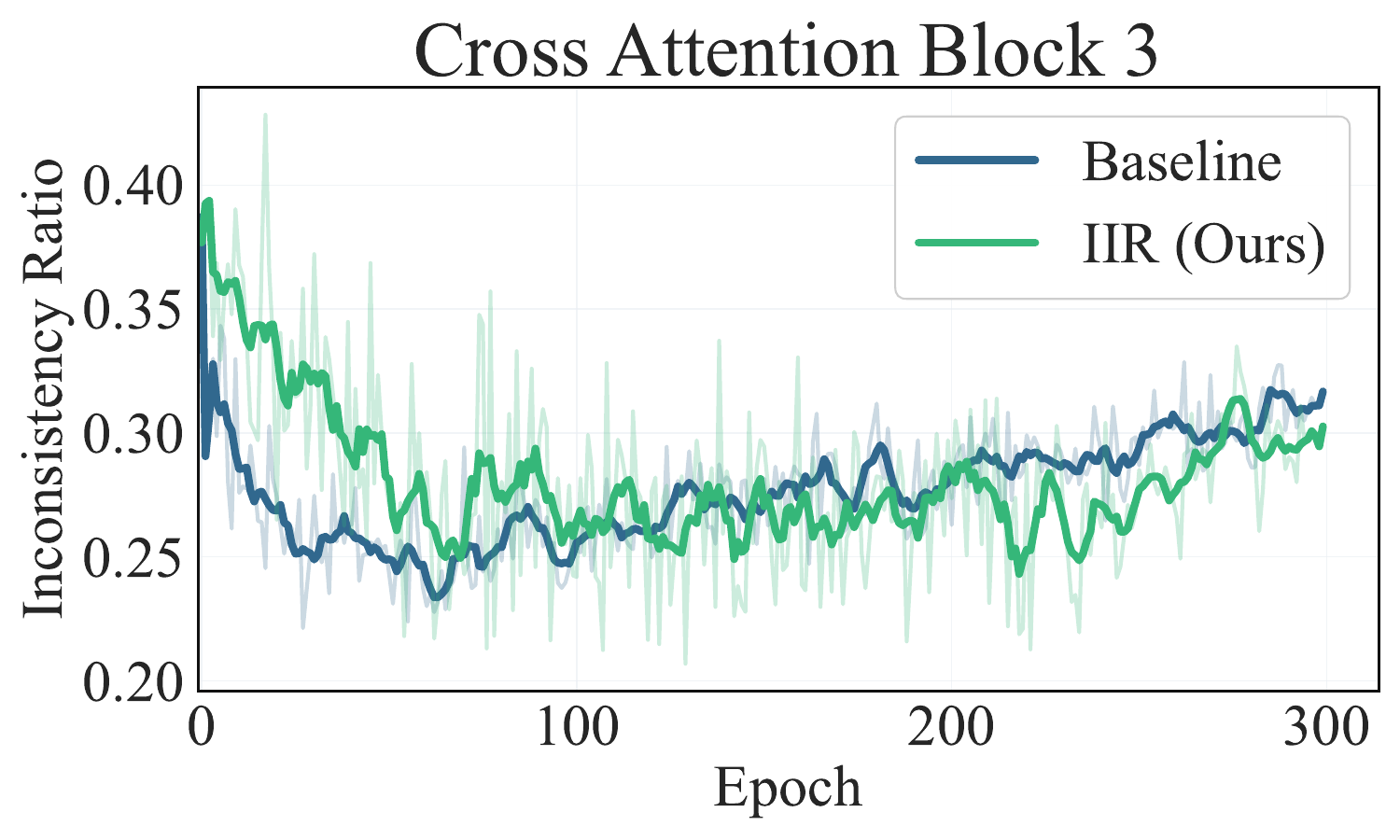}
    \label{subfig:CAB3}
  }
  \hfill % 第二行两子图之间的间距
  % 第二行第二列子图
  \subfigure[The inconsistency ratio of CAB 4]{
    \includegraphics[width=0.225\textwidth]{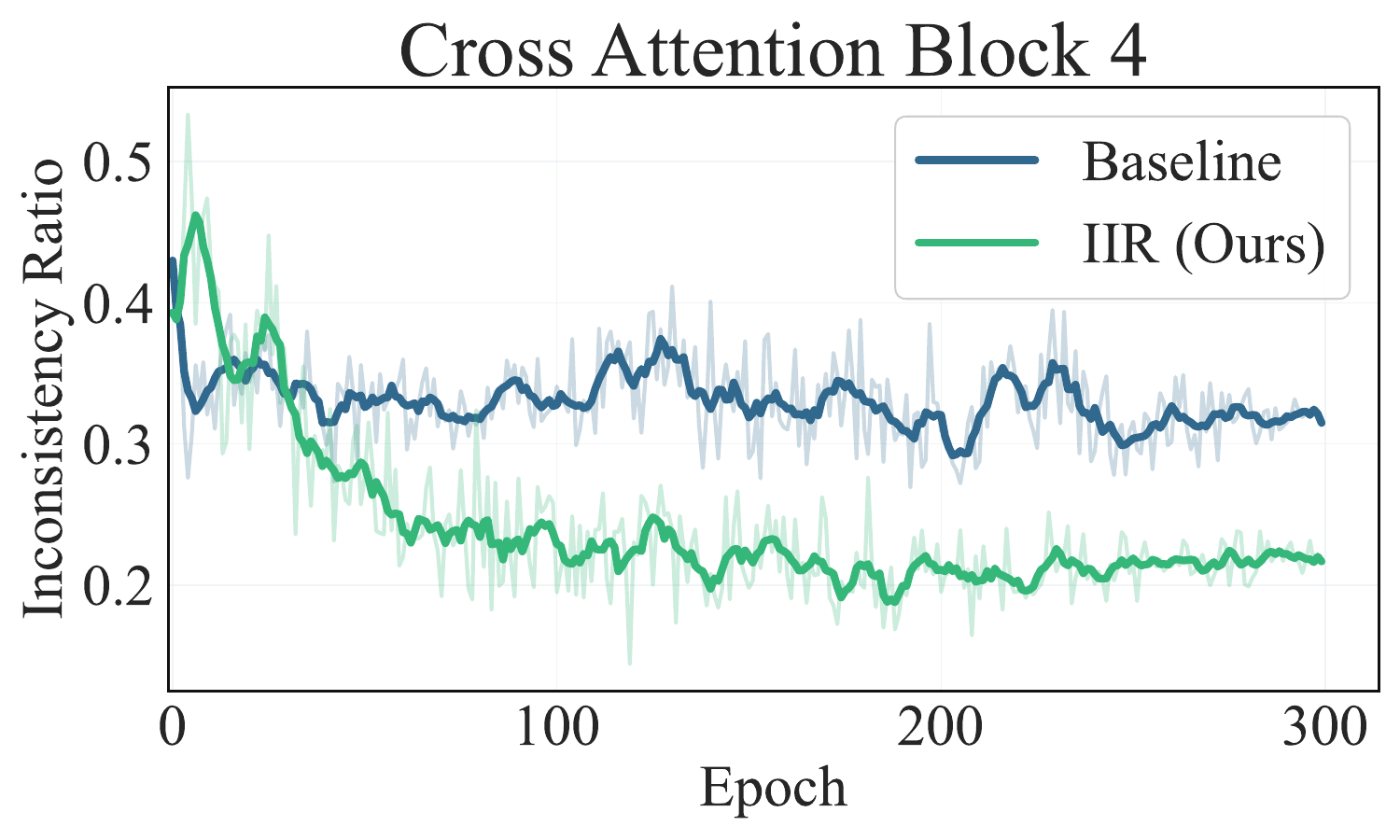}
    \label{subfig:CAB4}
  }
  % 总标题（描述4张子图的整体内容）
  \caption{Comparison of the inconsistency ratio of four Cross Attention Blocks (CAB) during the training process.}
  \label{fig:CAB}
\end{figure}

\subsection{Directional Dropout}
\label{sec:motivation DD}
\textbf{Intra-Image Interference} causes support-query inconsistency during training. 
The performance of the cross attention decoder can be evaluated by the inconsistency ratio between support and query features, which is defined as the proportion of support-query pairs in the support-query cross attention map where query feature $i$ and its most similar support feature $j$ do not satisfy $M_i=M_j$. In our model, the inconsistency ratio is mask ratio $r$ times $h_q\times w_q$.
The mask pairs of query feature $i$ and its most similar support feature $j$ should be consistent, meaning that \textit{the inconsistency ratio should decrease during training}, as shown in Fig. \ref{fig:CAB}. Comparison between our method and the baseline reveals that our inconsistency ratio is lower than that of the baseline, demonstrating that our model can more effectively suppress the support-query feature inconsistency.

We further analyzed the inconsistency ratios of different methods.
Table \ref{tab:Mask ratio} demonstrates that the average support-query inconsistency ratio of DDM is $(54.31 + 16.8 + 19.54 + 30.27 + 21.68)/5 = 28.52\%$, which represents a reduction of 2.4\% compared with CyCTR’s $(48.87 + 27.76 + 22.26 + 32.36 + 23.38)/5 = 30.93\%$. 
One potential reason is that, compared to CyCTR’s ``column masking" strategy, DDM eliminates approximately $30.93\%\times 899/900=30.90\%$ of information loss.
This verifies that DDM’s ``directional masked inconsistent feature pair" strategy can accurately suppress false positives and false negatives, rather than excessively masking valid information as CyCTR does. Consequently, DDM alone contributes a 0.9\% mIoU improvement, and achieves further performance enhancement when synergized with PMGM.
\begin{table}[t]
\centering
\caption{Support-Query inconsistency ratio of baseline, CyCTR and DDM across Cross Attention Blocks (CAB) and folds during testing, in \%.}
\resizebox{\linewidth}{!}{
\begin{tabular}{cc|ccccc}
\hline
 & & Fold-0 & Fold-1 & Fold-2 & Fold-3 & Mean \\ \hline
\multirow{5}{*}{Baseline} & CAB 0 & 29.37 & 27.84 & 27.95 & 27.78 & 28.24 \\
  & CAB 1 & 29.78 & 26.38 & 17.71 & 17.38 & 22.81 \\
  & CAB 2 & 20.98 & 29.51 & 39.85 & 11.60 & 25.49 \\
  & CAB 3 & 38.27 & 28.31 & 21.14 & 38.98 & 31.68 \\
  & CAB 4 & 37.71 & 22.82 & 55.58 & 9.83  & 31.49 \\ \hline
\multirow{5}{*}{CyCTR} & CAB 0 & 22.37 & 24.57 & 91.13 & 57.39 & 48.87 \\
  & CAB 1 & 44.51  & 24.12 & 24.17 & 18.23 & 27.76 \\
  & CAB 2 & 19.20 & 23.32 & 21.26 & 25.27 & 22.26 \\
  & CAB 3 & 20.22 & 23.62 & 24.54 & 61.05 & 32.36 \\
  & CAB 4 & 20.59 & 23.33 & 24.59 & 25.01 & 23.38 \\ \hline
\multirow{5}{*}{IIR}  
  & CAB 0 & 54.90 & 48.20 & 58.91 & 55.24 & 54.31 \\
  & CAB 1 & 7.35  & 11.46 & 31.09 & 17.30 & 16.80 \\
  & CAB 2 & 14.57 & 24.44 & 14.27 & 24.88 & 19.54 \\
  & CAB 3 & 23.56 & 30.74 & 55.55 & 11.22 & 30.27 \\
  & CAB 4 & 32.02 & 31.53 & 12.20 & 10.98 & 21.68 \\ \hline
\end{tabular}}
\label{tab:Mask ratio}
\end{table}

\subsection{Limitation}
The first key improvement of IIR is extending a single prototype to multiple ones, thereby more accurately comparing the support foreground with query images and obtaining a more precise query prior map.
CAM is a method that is easy to implement and proven to excel at localizing the most salient foreground regions in images, and it has successfully validated the effectiveness of our method's theory. However, CAM also introduces the burden of additionally training a classification head as well as computational overhead from backpropagation; future work may consider developing more efficient and reliable methods for foreground region localization.
Besides, existing FSS methods, including our proposed IIR, fail to achieve foreground segmentation with clinical utility when confront with real-world medical scenarios such as fundus vessels segmentation by scarce data and complex foreground topological structures. Future research needs to further explore approaches to enhance the performance of FSS methods on such targets.
\section{Conclusion}
\label{sec:conclusion}
In this work, we present IIR model, which refines the prior map and cross attention map generation processes to solve Inter- and Intra-image Gap.
The first key component is the PMGM, which generates more accurate prior map by using CAM and categorical consistency assumption, bridging the inter support-query gap.
The second key component is the DDM, an architecture that is designed to effectively identify specific inconsistencies in feature extraction, enabling the model to mask specific erroneous support-query pixel pairs, benefiting decoder blocks during training.
Extensive experiments and ablation studies demonstrate that IIR achieves new SOTA results on 9 benchmarks, surpassing most of previous methods in both 1-shot and 5-shot segmentation settings. By integrating PMGM and DDM, our model effectively mitigates both inter- and intra-image discrepancies, resulting in more robust segmentation across diverse object categories. 
\newpage

\bibliographystyle{IEEEtran}
\bibliography{main}

\clearpage
\setcounter{page}{1}

\appendix   
\setcounter{table}{0}   %从0开始编号，显示出来表会A1开始编号
\setcounter{figure}{0}
\setcounter{equation}{0}
\setcounter{algocf}{0}
\renewcommand{\thetable}{S\arabic{table}}
\renewcommand{\thealgocf}{S\arabic{algocf}}
\renewcommand{\thefigure}{S\arabic{figure}}
\renewcommand{\theequation}{S\arabic{equation}}

\subsection{CyCTR Method}
\label{sec:supp method}
CyCTR\cite{CyCTR} considers the cycle consistency between the support and query features: if pixel $j$ in the support features is most similar to pixel $max_q$ in the query features, and pixel $max_q$ in the query features is most similar to pixel $max_s$ in the support features, then the masks of $j$ and $max_s$ should be consistent, as shown in Algorithm \ref{al:supp CyCTR}. We also present our DDM algorithm in Algorithm \ref{al:supp DDM} for better comparison.
\begin{algorithm}
\caption{CyCTR\label{al:supp CyCTR}}
\renewcommand{\KwData}{\textbf{Input:}}
\renewcommand{\KwResult}{\textbf{Output:}}
\KwData{\\cross attention map $A\in\mathbb{R}^{h_qw_q\times h_sw_s}$, \\flatten support mask $M^s\in\mathbb{R}^{h_sw_s}$, \\cross attention mask$A^s\in\mathbb{R}^{h_qw_q\times h_sw_s}$ initialized by 0}
\KwResult{cross attention map $A\in\mathbb{R}^{h_qw_q\times h_sw_s}$}

\For{$j=1\ to\ h_sw_s$}{
$max_q=\arg\max_i A[i,j]$\;
$max_s=\arg\max_k A[max_q,k]$\;
\If{$M^s[j]\neq M^s[max_s]$}{$A^s[:,j]=-\infty$\;}
}
$A=A+A^s$\;
\end{algorithm}

\begin{algorithm}
\caption{Directional Dropout Module}\label{al:supp DDM}
\renewcommand{\KwData}{\textbf{Input:}}
\renewcommand{\KwResult}{\textbf{Output:}}
\KwData{\\cross attention map $A\in\mathbb{R}^{h_qw_q\times h_sw_s}$, \\flatten support mask $M^s\in\mathbb{R}^{h_sw_s}$, \\flatten query mask $M^q\in\mathbb{R}^{h_qw_q}$}

\KwResult{masked cross attention map $A'\in\mathbb{R}^{h_qw_q\times h_sw_s}$}

Initialize cross attention mask $A^s\in\mathbb{R}^{h_qw_q\times h_sw_s}$ and mask ratio $r$ with 0\;
\For{$i=1\ to\ h_qw_q$}{
$j=\arg\max_{k \in \{1,2,...,h_sw_s\}} A[i,k]$\;
\If{$M^q[i]\neq M^s[j]$}{$A^s[i,j]=-\infty$\;$r=r+\frac{1}{h_qw_q\times h_sw_s}$\;}
}
$A'=A+A^s$\;
$A'=A'\times\frac{1}{1-r}$\;
\end{algorithm}

\subsection{Datasets}
\label{sec:dataset}
We select a total of 9 existing and 2 new few shot datasets across three tasks for experimental validation:
\begin{enumerate}
 \item PASCAL-5$^i$ \cite{SG-ONE}: \textit{FSS.} The PASCAL-5$^i$ consists of PASCAL VOC 2012 dataset \cite{PASCAL} and the SBD dataset \cite{SBD}. The 20 classes in PASCAL-5$^i$ are divided into 4 folds, with each fold containing 5 classes. During training, we use 3 fold for training and reserve 1 fold for testing, following SG-ONE \cite{SG-ONE}.
 \item COCO-20$^i$ \cite{SG-ONE}: \textit{FSS.} COCO-20$^i$ is originated from COCO \cite{COCO}, the data are split into 4 folds, with each fold containing 20 classes.
 \item LVIS-92$^i$ \cite{Matcher}: \textit{FSS.} LVIS-92$^i$ is based on the LVIS dataset \cite{LVIS}, a more challenging benchmark for evaluating the generalization of a model across datasets. After removing the classes with less than two images, the dataset retain 920 classes for training and testing. These classes were then divided into 10 equal folds.
 \item PACSCAL-Part \cite{Matcher}: \textit{OPFSS.} The dataset is built based on PASCAL \cite{PASCAL} and its body part annotations \cite{pascal_part}. The dataset consists of four superclasses: animals, indoor, person, and vehicles.
 \item PACO-Part \cite{Matcher}: \textit{OPFSS.} The dataset is built based on PACO dataset \cite{PACO} with 303 object part categories. The dataset consists four folds, each with about 76 object parts.
 \item FSS-1000 \cite{FSS1000}: \textit{CDFSS (natural → natural).} The dataset comprising 1000 natural image categories, with each category contains 10 samples.
 \item Deepglobe \cite{deepglobe}:  \textit{CDFSS(natural → remote sensing).} The dataset consists of satellite images with 7 foreground classes: urban, agriculture, rangeland, forest, water, barren, and unknown.
 \item ISIC2018 \cite{isic}: \textit{CDFSS (natural → medical).} The dataset comprises 2,596 images of skin cancer screening, with each image depicting exactly one primary lesion.
 \item Chest X-ray \cite{chest1,chest2}: \textit{CDFSS (natural → medical).} Dataset consists X-ray Tuberculosis images. It includes 566 images, which are collected from 58 cases with a manifestation of Tuberculosis and 80 normal cases.
\item Fives: \textit{CDFSS (natural → medical).} The FIVES dataset \cite{fives} consists of 800 color fundus photographs with 2048×2048 multi-disease pixel-wise retinal vessel annotation, including 200 healthy, 200 age related macular degeneration, 200 diabetic retinopathy, and 200 glaucoma. To the best of our knowledge, this is the largest retinal vessel segmentation dataset. 
\item Stare: \textit{CDFSS (natural → medical) few shot semantic segmentation.} The STARE dataset \cite{stare} contains 20 fundus images with a resolution of 700x605 pixels. Images cover various lesion types, such as macular degeneration, hypertensive retinopathy, and diabetic retinopathy, among others.
\end{enumerate}

\subsection{Implementation Details}
\label{sec:Implementation}
The implementation details for each dataset follow their respective baselines, as specified below:
\begin{enumerate}
    \item PASCAL-5$^i$ and COCO-20$^i$: Our model adopts BAM \cite{BAM} as its baseline. Since the first stage of BAM fine-tunes the encoder model, we incorporate the KL divergence loss into the classification head to maintain the consistency of class mapping, which ensures that CAM can correctly map to foreground regions. During the inference phase, the Directional Dropout Module is not employed due to the unknown query mask. For this study, ResNet50 and ResNet101 \cite{ResNet} are used as backbones. For the PASCAL-5$^i$ dataset, the image size, batch size, and learning rate are set to 473×473, 8, and 0.05, respectively. For COCO-20$^i$, these parameters are configured as 641×641, 8, and 0.05, respectively.
    \item LVIS-92$^i$: We follow the setting of Matcher \cite{Matcher}, which uses the model trained on COCO-20$^i$ for testing directly.
    \item PASCAL-part and PACO-Part: Following the experimental settings of PerSAM \cite{PerSAM} and Matcher \cite{Matcher}, we use DINOv2 \cite{dinov2} (with ViT-L/14 \cite{vit}) as the feature extraction backbone. For the PASCAL-Part dataset, the model is first trained on the PASCAL-5$^i$ and then transferred to PASCAL-Part; For the PACO-Part dataset, the model is first trained on the COCO-20$^i$ and then transferred to PACO-Part. For both datasets, the image size is set to 518×518, batch size to 4.
    \item Cross domain few shot datasets: Our model takes IFA \cite {IFA} as its baseline. The initial model is first trained on PASCAL-5$^i$ and then fine-tuned on a cross-domain dataset following the second stage of IFA. During this fine-tuning stage, the augmented support image-mask pair is utilized as the query image-mask pair for domain adaptation. Finally, the model is tested on the specific dataset. We adopt ResNet50 \cite{ResNet} as the backbone of our model. Following the baseline IFA \cite{IFA}, the size of both the support and query images is set to 400×400. During the fine-tuning step, the SGD optimizer is used and the learning rate is set to 0.01. The weight decay is set to 1e-2 for ISIC and 1e-3 for other datasets. Since the pixel-wise retinal vessel segmentation task has the problem of unbalanced foreground and background, the AdamW optimizer with a learning rate of 5e-4 and the weighted BCE loss are employed.
\end{enumerate}

\subsection{More Ablation Study}
\label{subsec:more ablation}
\begin{table}[t]
\centering
\caption{Performance of using different the number of Cross Decoder blocks and Decoder blocks, where CAB is the abbreviation of \textbf{C}ross \textbf{A}ttention \textbf{B}lock. (The Cross Attention Block number must be less than Decoder number)}
\begin{tabular}{c|ccccc}
\hline
\diagbox{CAB}{mIoU}{Decoder} & 2 & 3 & 4 & 5 & 6 \\ \hline
0 & 70.87 & 71.49 & 71.59 & 71.61 & 71.53 \\
1 & 71.03 & 71.57 & 71.61 & 71.83 & 71.95  \\
2 & 71.55 & 71.69 & 71.89 & 72.04 & 72.18  \\
3 & -     & 71.82 & 71.83 & 72.23 & 72.31  \\
4 & -     & -     & 72.28 & 72.67 & 72.49  \\
5 & -     & -     & -     & 72.45 & \textbf{72.91}  \\ 
6 & -     & -     & -     & -     & 72.53  \\ 
\hline
\end{tabular}
\label{table:Decoder Number}
\end{table}

\begin{figure}[htbp]
  \centering
  \includegraphics[width=\linewidth]{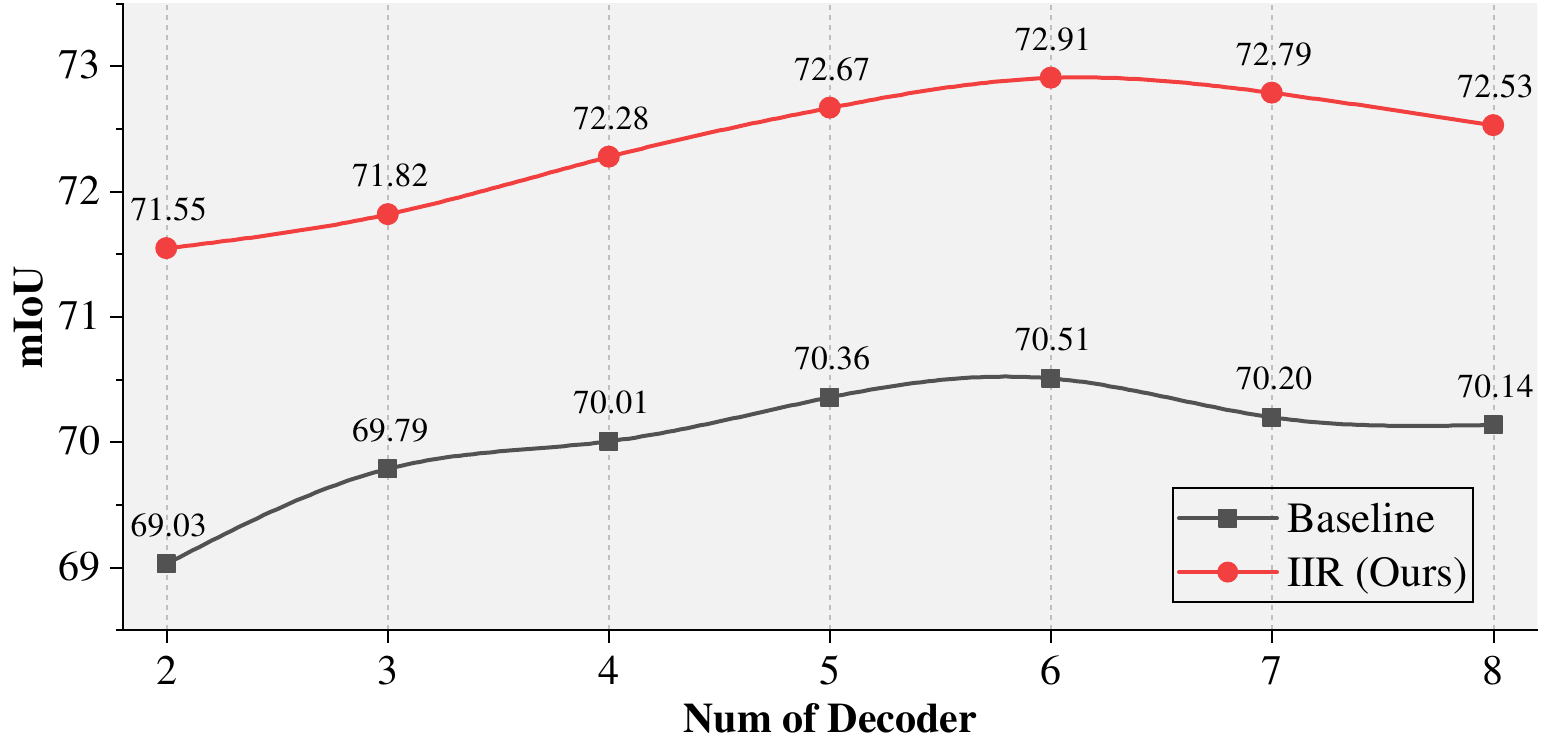}
  \caption{Saturation experiments on the number of decoders.}
  \label{fig:Saturation study}
\end{figure}

\subsubsection{Decoder Number}
The decoder number is a key factor affecting the results.
Table \ref{table:Decoder Number} and Fig. \ref{fig:Saturation study} demonstrate that when the number of decoders is 6 and the number of CABs is 5, the mIoU reaches 72.91\%. If the number of CABs is less than 5, the support-query feature interaction is insufficient (the mIoU is only 72.18\% when the number of CABs is 2). If the number of decoders exceeds 6, the performance enters a saturation stage (the mIoU drops to 72.53\% when the number of decoders is 7) with increased computational cost. This verifies that the architecture of IIR achieves an optimal balance between feature interaction and computational efficiency.

\begin{figure}[htbp]
  \centering
  \includegraphics[width=\linewidth]{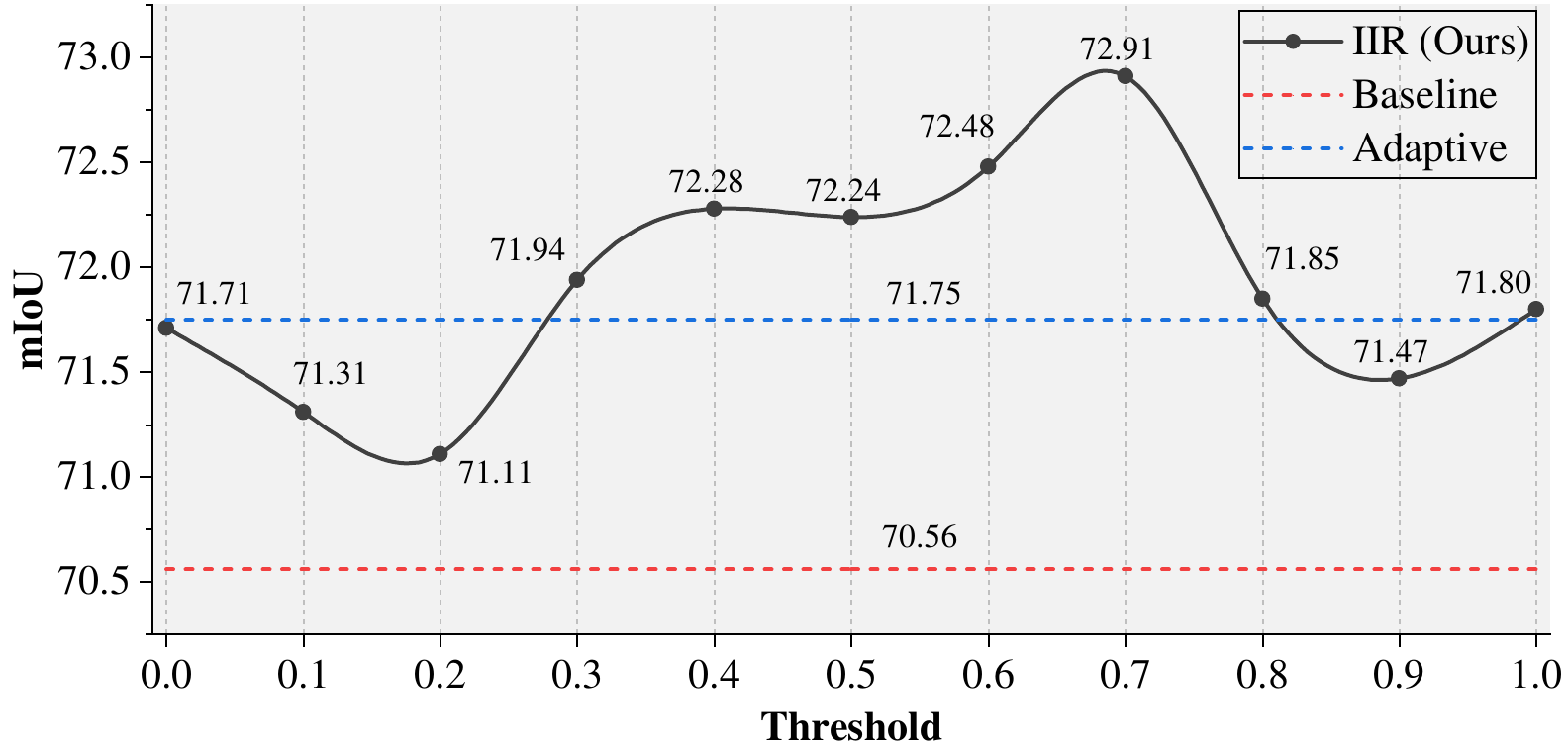}
  \caption{Performance of using different CAM threshold.}
  \label{fig:CAM threshold}
\end{figure}

\subsubsection{CAM Methods \& Thresholds}
Different CAM methods and thresholds focus on different aspects of images, and thus affect the prototype generation regions, thereby influencing the accuracy of prior maps and final prediction.
Table \ref{table:cam method} demonstrates that XGradCAM achieves the highest mIoU of 72.91\% among all CAM methods, outperforming methods such as GradCAM, as it can more accurately localize class-relevant regions. Fig. \ref{fig:CAM threshold} further confirms that the optimal performance is achieved when the CAM threshold is set to 0.7: a threshold below 0.3 leads to scattered foreground in the CAM, confusing foreground and background, while a threshold above 0.7 results in the loss of local features (e.g., target edge details). This component selection ensures that the dual prototypes generated by PMGM possess both global semantics and local specificity.
The adaptive threshold is defined within the range [0, 1] with an interval of 0.01. Specifically, it refers to the value that maximizes the mIoU between the corresponding support pseudo mask and the support ground truth mask.
We found that the adaptive threshold achieves a mIoU of 71.75\%, with the optimal thresholds typically falling within the range of [0.2, 0.5].
This limited performance is due to the fact that thresholds within this range leads to decentralization of the CAM foreground, causing confusion between foreground and background features. 

\begin{table}[htbp]
\centering
\caption{Performance of using different CAM method}
\begin{tabular}{c|c}
\hline
CAM Method& mIoU\\ \hline
XGradCAM(2020)\cite{XGradCAM} & \textbf{72.91}  \\
HiResCAM(2021)\cite{HiResCAM} & 71.91   \\
GradCAM(2017)\cite{GradCAM} & 72.23 \\
BagCAM(2023)\cite{bagcam}& 71.54    \\
GradCAM++(2018)\cite{GradCAM++} & 71.67 \\ \hline
\end{tabular}
\label{table:cam method}
\end{table}

\subsection{More Visualization}
We provide supplementary visualizations of the PASCAL-5$^i$, LVIS-92$^i$, and paco-part datasets.

\begin{figure*}[htbp]
  \centering
  \includegraphics[width=0.95\textwidth]{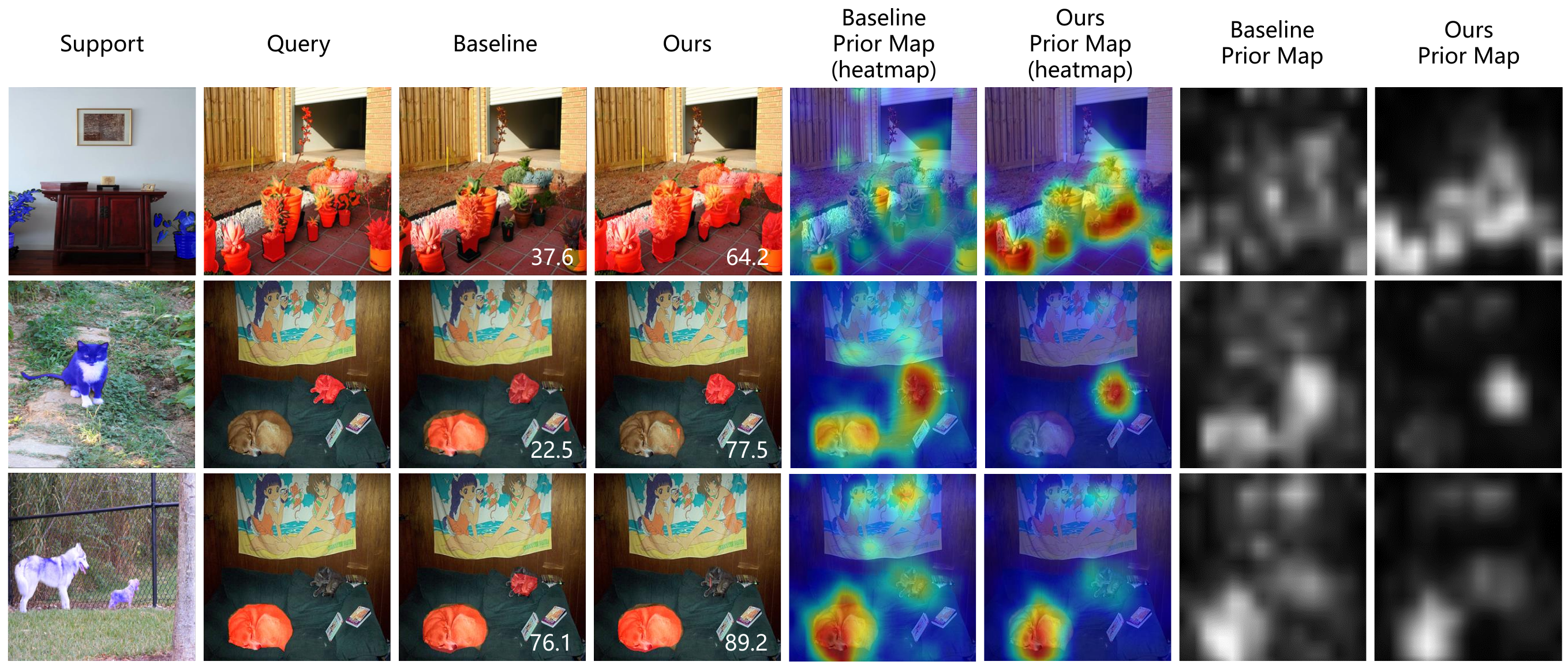}
  \caption{More visualization on PASCAL-5$^i$ dataset.}
  \label{fig:vis more pascal}
\end{figure*}

\begin{figure}[htbp]
  \centering
  \includegraphics[width=0.95\linewidth]{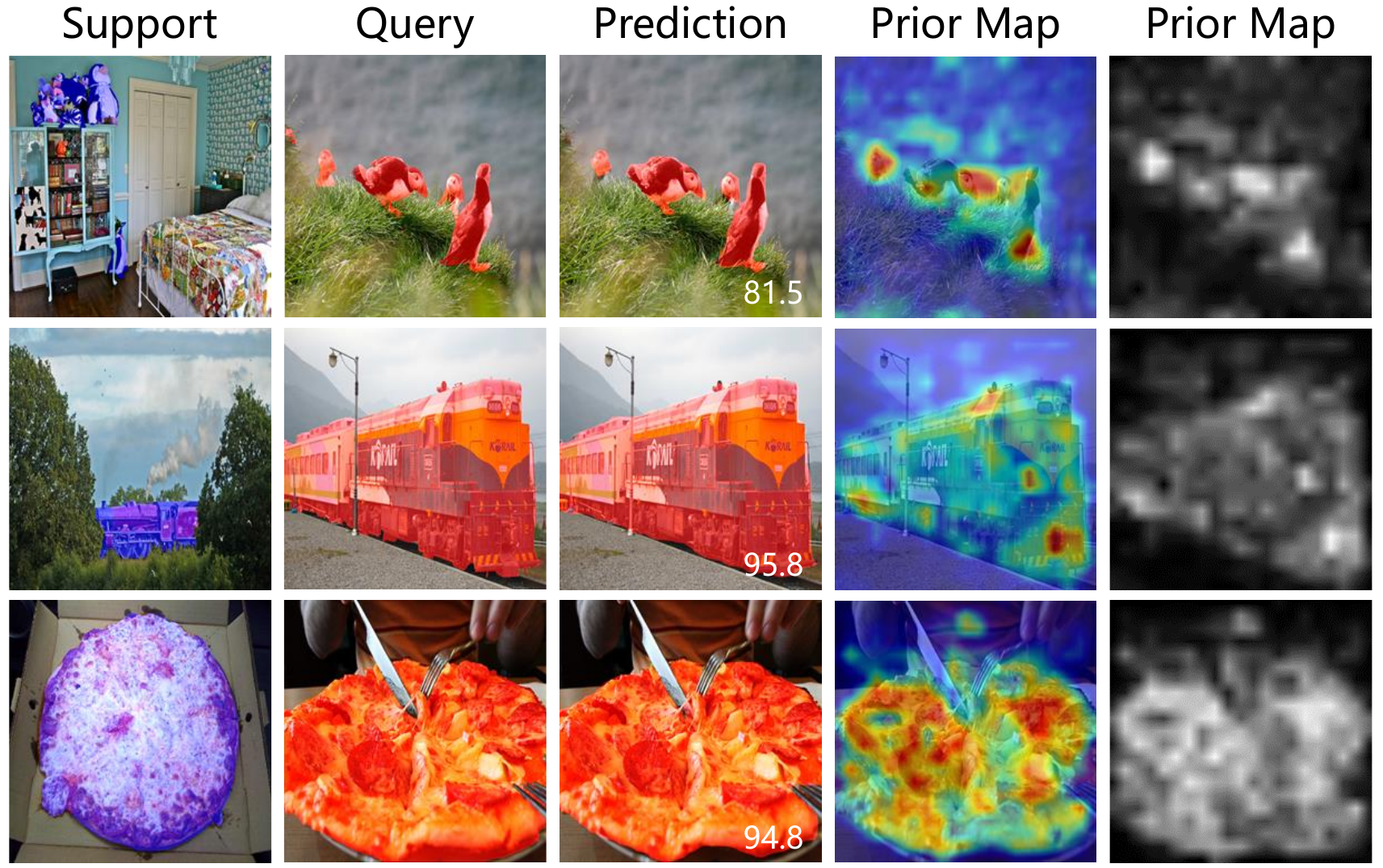}
  \caption{Visualization on LVIS-92$^i$ dataset for multiple targets, size-varying targets, and irregularly shaped targets.}
  \label{fig:lvis}
\end{figure}

\begin{figure}[t]
  \centering
  \includegraphics[width=0.95\linewidth]{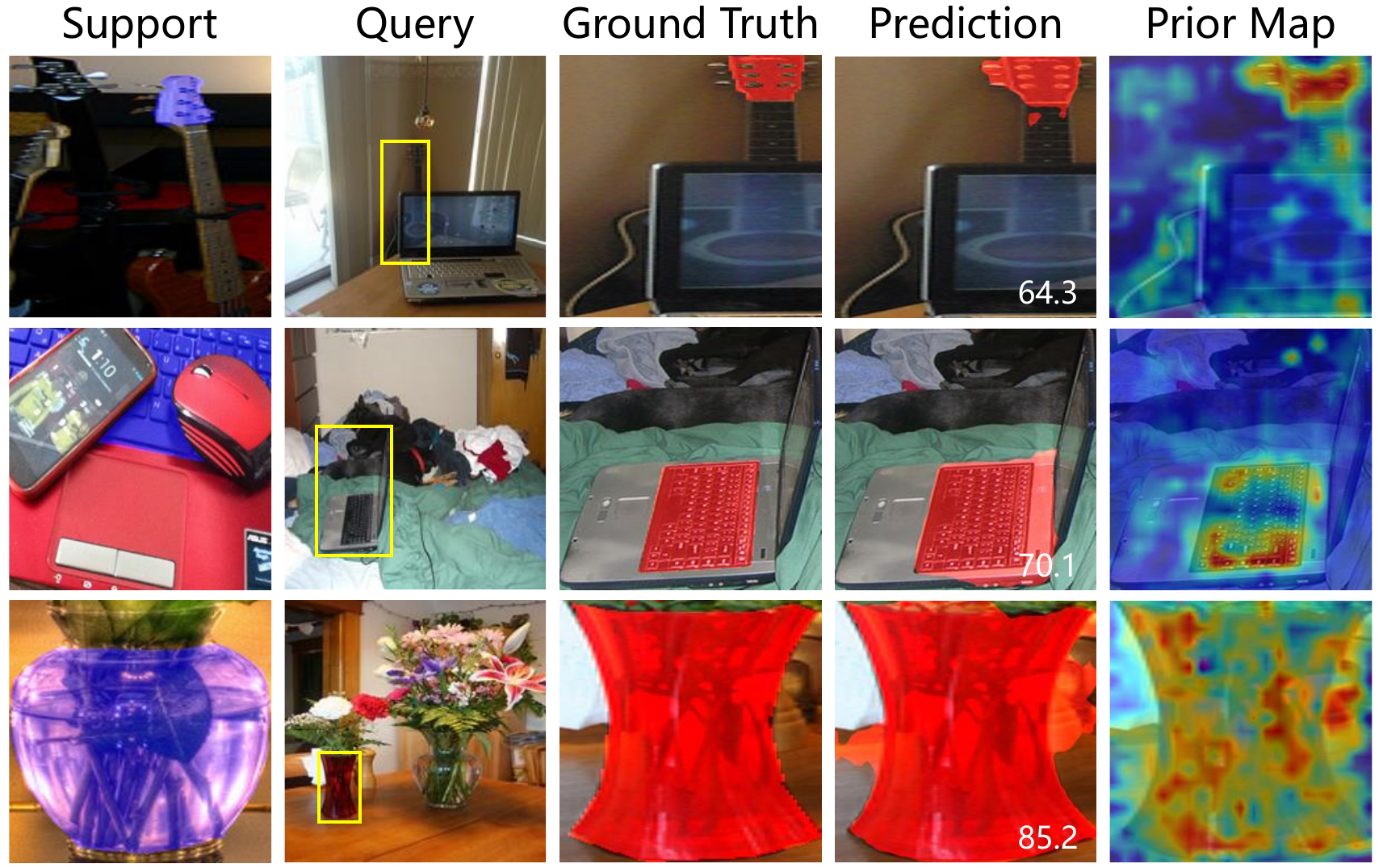}
  \caption{Visualization on PACO-Part dataset. The yellow box in the query image (second column) serves as model input (third column).}
  \label{fig:paco-part}
\end{figure}

Additional demonstrations for the PASCAL-5$^i$ dataset under the ResNet50 1-shot configuration are presented in Fig. \ref{fig:vis more pascal}. We have additionally conducted corresponding visualizations on the LVIS-92$^i$ dataset, which are displayed in Fig. \ref{fig:lvis}. These results illustrate that our approach achieves superior segmentation performance for multiple targets, targets of varying sizes, and irregularly shaped targets. Fig. \ref{fig:paco-part} showcases several representative segmentation results to depict typical scenarios in the few shot object part segmentation task. In this task, the model is required to perform segmentation on cropped images, as illustrated in the second column of the figure.

\subsection{Error Cases Analysis}
\begin{figure}[htbp]
  \centering
  \includegraphics[width=\linewidth]{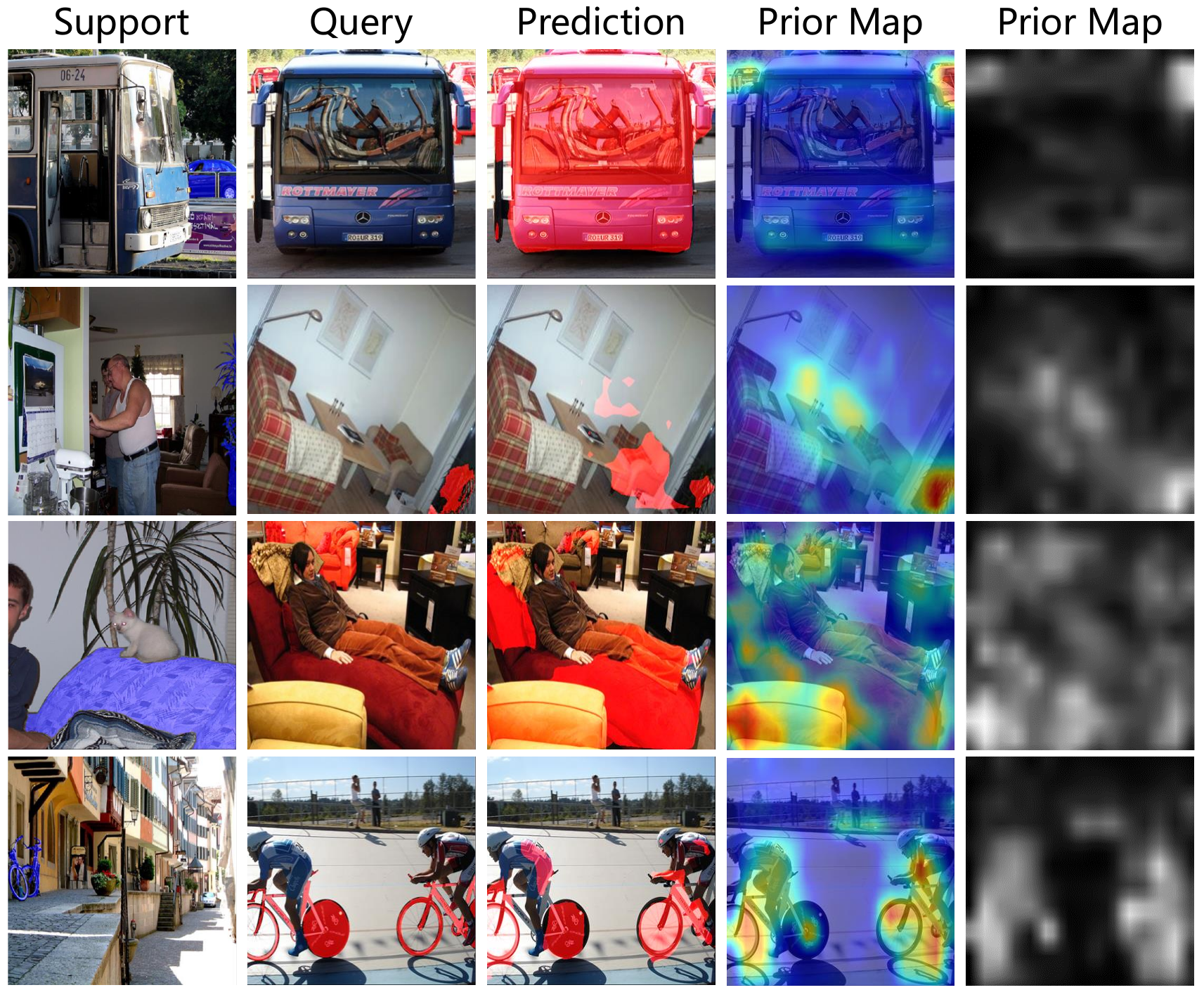}
  \caption{Visualization of error cases on PASCAL-5$^i$ dataset.}
  \label{fig:vis error pascal}
\end{figure}

\begin{figure}[htbp]
  \centering
  \includegraphics[width=\linewidth]{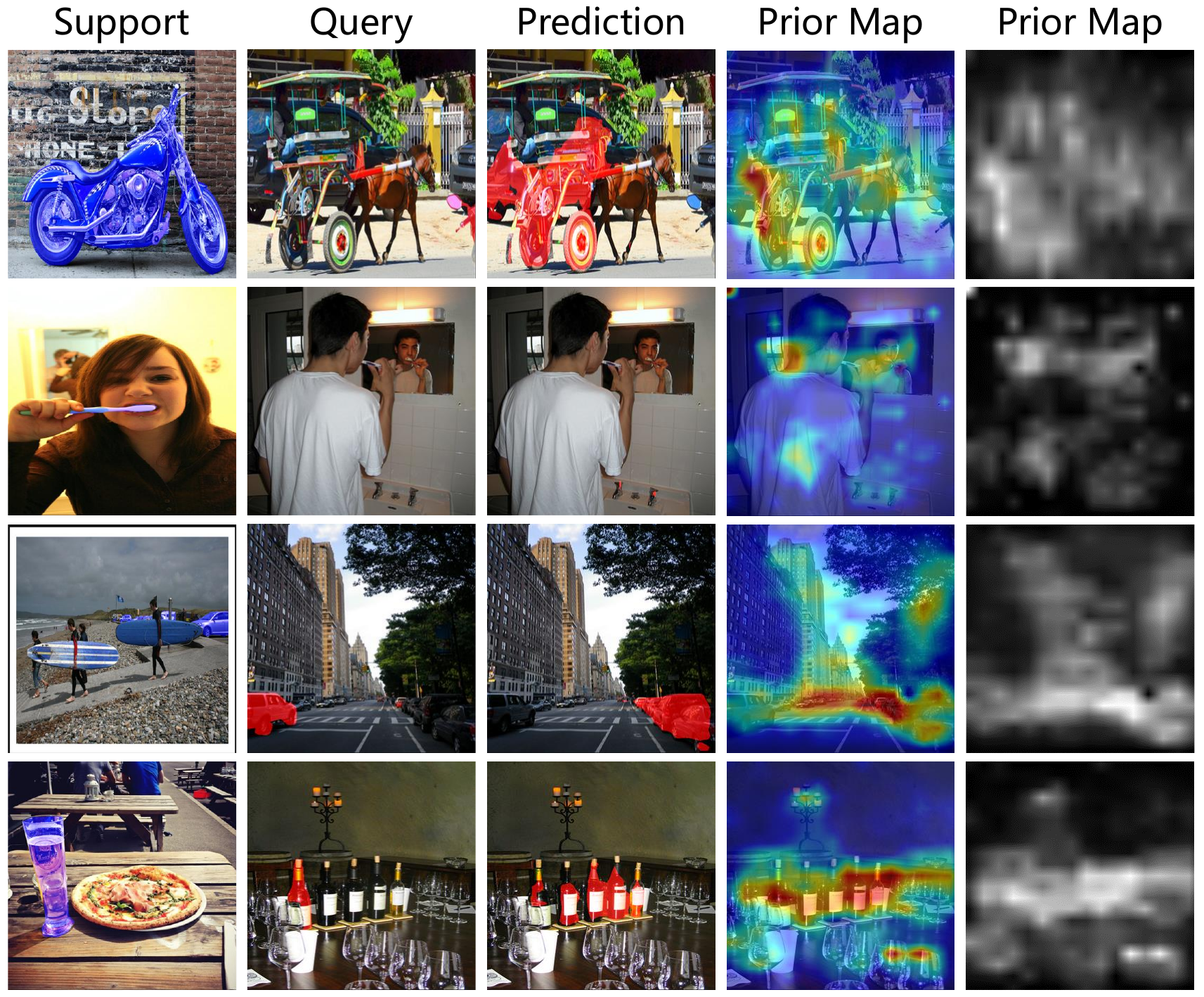}
  \caption{Visualization of error cases on COCO-20$^i$ dataset.}
  \label{fig:vis error coco}
\end{figure}

We also investigate several error cases in the datasets. These cases can can be categorized as follows:
\begin{itemize}
    \item Small foreground hidden behind large background objects: when the ground truth mask is obscured by a large object in the background, the model tends to misclassify that large object as the foreground. (e.g. the first row in Fig. \ref{fig:vis error pascal} and \ref{fig:vis error coco})
    \item Camouflaged target or small target segmentation. We notice that some foreground objects has merge with the background or are different to discern in query or support images, leading the model to fail in extracting useful features from them (e.g. the second row in Fig. \ref{fig:vis error pascal} and \ref{fig:vis error coco}).
    \item Objects with similar features are classified into different classes. We observe that certain minority classes are difficult to distinguish, particularly in the COCO-20$^i$ dataset which contains more classes to be segmented (e.g. the third row in Fig. \ref{fig:vis error coco}).
    \item Errors in mask labeling: we find that some annotations are incomplete (e.g. the third row in Fig. \ref{fig:vis error pascal} and the last row in Fig. \ref{fig:vis error coco}) and certain class is annotated with foreground contours (e.g. the `bicycle' class in PASCAL-5$^i$, as shown in the last row in Fig. \ref{fig:vis error pascal}).
\end{itemize}

\end{document}